\documentclass[10pt,twocolumn,letterpaper]{article}

\usepackage[pagenumbers]{cvpr} 

\usepackage{graphicx}
\usepackage{amsmath}
\usepackage{amssymb}
\usepackage{booktabs}
\usepackage{multirow}
\usepackage{soul}
\usepackage{color}
\usepackage{xcolor}

\usepackage{adjustbox}
\usepackage{makecell}
\usepackage{bbding}

\usepackage[pagebackref,breaklinks,colorlinks]{hyperref}

\usepackage[capitalize]{cleveref}
\crefname{section}{Sec.}{Secs.}
\Crefname{section}{Section}{Sections}
\Crefname{table}{Table}{Tables}
\crefname{table}{Tab.}{Tabs.}
\newcommand\blfootnote[1]{%
  \begingroup
  \renewcommand\thefootnote{}\footnote{#1}%
  \addtocounter{footnote}{-1}%
  \endgroup
}

\def\cvprPaperID{1812} 
\def\confName{CVPR}
\def\confYear{2023}

\begin{document}

\title{Prompt, Generate, then Cache:\\Cascade of Foundation Models makes Strong Few-shot Learners}

\author{Renrui Zhang$^{*2,3}$, Xiangfei Hu$^{*2,4}$, Bohao Li$^{5}$, Siyuan Huang$^{2,4}$, Hanqiu Deng$^{2}$,\\
Hongsheng Li$^{3}$, Yu Qiao$^{2}$, Peng Gao$^{\dagger1,2}$ \vspace{0.2cm}\\
  $^1$Shenzhen Institutes of Advanced Technology, Chinese Academy of Science\\
  $^2$Shanghai Artificial Intelligence Laboratory
 \quad 
  $^3$The Chinese University of Hong Kong \quad\\
  $^4$Shanghai Jiaotong University \quad
  $^5$University of Chinese Academy of Sciences\vspace{0.2cm}\\
\texttt{\{zhangrenrui, huangsiyuan, qiaoyu, gaopeng\}@pjlab.org.cn},\\
\texttt{sjtuhxf@sjtu.edu.cn},\quad \texttt{hsli@ee.cuhk.edu.hk}
}
\maketitle
\newcommand{\hsy}[1]{{\color{blue}{#1}}}
\blfootnote{$^*$ Equal contribution.\ $\dagger$ Corresponding author}
\begin{abstract}
Visual recognition in low-data regimes requires deep neural networks to learn generalized representations from limited training samples. Recently, CLIP-based methods have shown promising few-shot performance benefited from the contrastive language-image pre-training. We then question, if the more diverse pre-training knowledge can be cascaded to further assist few-shot representation learning. In this paper, we propose \textbf{CaFo}, a \textbf{Ca}scade of \textbf{Fo}undation models that incorporates diverse prior knowledge of various pre-training paradigms for better few-shot learning. Our CaFo incorporates CLIP's language-contrastive knowledge, DINO's vision-contrastive knowledge, DALL-E's vision-generative knowledge, and GPT-3's language-generative knowledge. Specifically, CaFo works by `Prompt, Generate, then Cache'. Firstly, we leverage GPT-3 to produce textual inputs for prompting CLIP with rich downstream linguistic semantics. Then, we generate synthetic images via DALL-E to expand the few-shot training data without any manpower. At last, we introduce a learnable cache model to adaptively blend the predictions from CLIP and DINO. By such collaboration, CaFo can fully unleash the potential of different pre-training methods and unify them to perform \textit{state-of-the-art} for few-shot classification. Code is available at \url{https://github.com/ZrrSkywalker/CaFo}.

\end{abstract}

\section{Introduction}
\label{sec:intro}

\begin{figure}[h]
  \centering
   \includegraphics[width=1\linewidth]{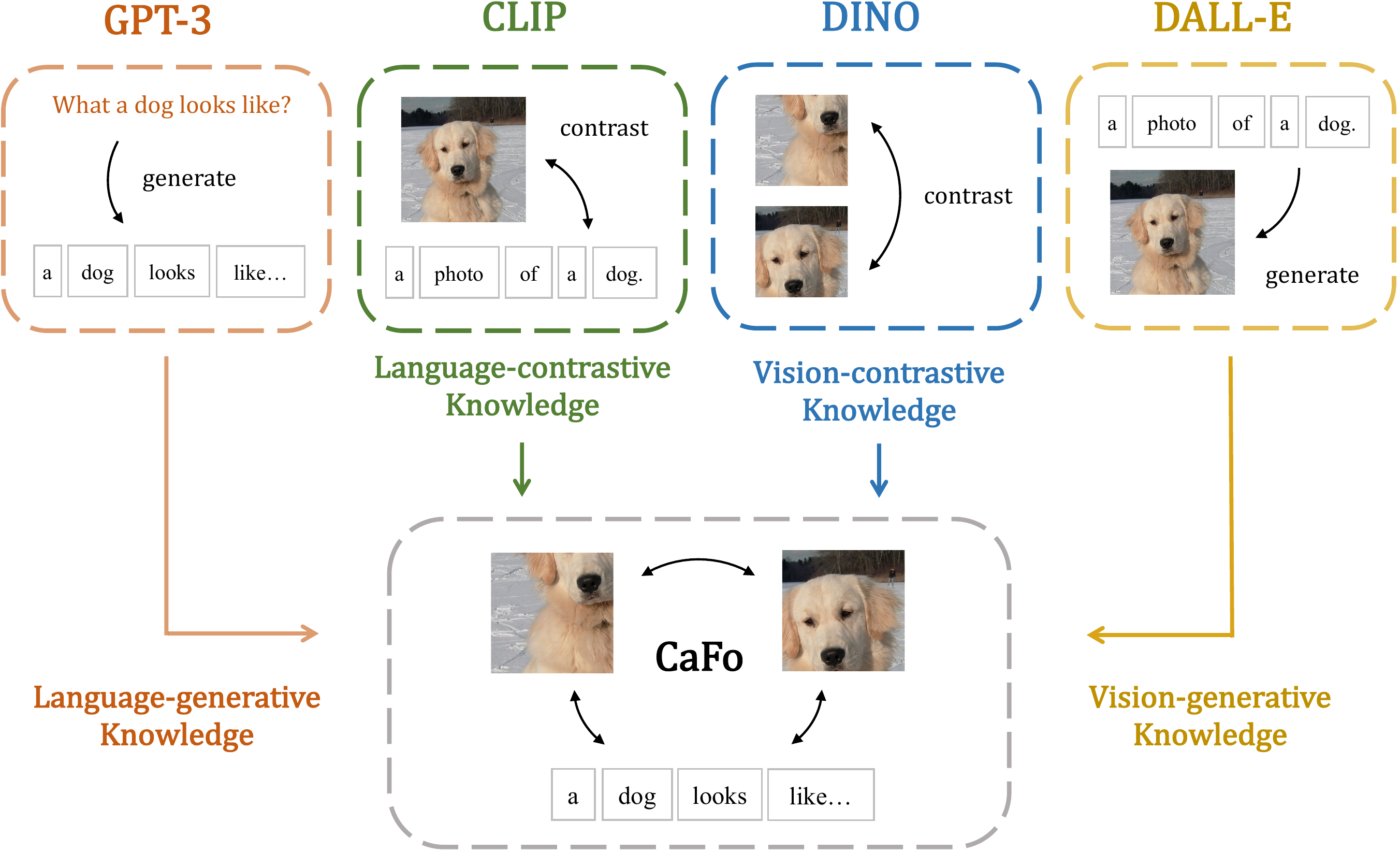}
   \caption{\textbf{The Cascade Paradigm of CaFo.} We adaptively incorporate the knowledge from four types of pre-training methods and achieve a strong few-shot learner.}
   \label{fig1}
\end{figure}
Convolutional neural networks~\cite{NIPS2012_c399862d} and transformers~\cite{NIPS2017_3f5ee243} have attained great success on a wide range of vision tasks with abundant datasets~\cite{deng2009imagenet}. Instead, for some data-deficient and resource-finite scenarios, few-shot learning~\cite{NIPS2016_90e13578, snell2017prototypical} also becomes a research hotspot, where the networks are constrained to learn from limited images with annotations. Many previous works have been proposed in this field to enhance model's generalization capability by meta learning~\cite{finn2017model, LearningToLearn16}, metric learning~\cite{zhang2020deepemd}, and data augmentation~\cite{Hallucinating17, yang2021free}. Recently, CLIP~\cite{radford2021learning} pre-trained by large-scale language-image pairs shows favorable zero-shot transfer ability for open-vocabulary visual recognition. The follow-up CoOp~\cite{coop}, CLIP-Adapter~\cite{2110.04544} and Tip-Adapter~\cite{zhang2021tip} further extend it for few-shot classification and achieve superior performance on various downstream datasets. This indicates that, even if the few-shot training data is insufficient, the large-scale pre-training has endowed the network with strong representation ability, which highly benefits the few-shot learning on downstream domains. Now that there exist various self-supervisory paradigms besides CLIP, \textit{could we adaptively integrate their pre-learned knowledge and collaborate them to be a better few-shot learner?}


To tackle this issue, we propose \textbf{CaFo}, a \textbf{Ca}scade of \textbf{Fo}oundation models blending the knowledge from multiple pre-training paradigms with a `Prompt, Generate, then Cache' pipeline. As shown in Figure~\ref{fig1}, we integrate CLIP~\cite{radford2021learning}, DINO~\cite{Caron_2021_ICCV}, DALL-E~\cite{pmlr-v139-ramesh21a}, and GPT-3~\cite{brown2020language} to provide four types of prior knowledge for CaFo. Therein, CLIP~\cite{radford2021learning} is pre-trained to produce paired features in the embedding space for every image and its descriptive text. Guided by texts with different categorical semantics, CLIP~\cite{radford2021learning} can well classify the images aided by \textbf{language-contrastive knowledge}. DINO follows contrastive self-supervised learning~\cite{Caron_2021_ICCV} to match the 
representations between two transformations of one same image, which is expert at distinguishing different images with \textbf{vision-contrastive knowledge}. Similar to CLIP~\cite{radford2021learning}, DALL-E~\cite{pmlr-v139-ramesh21a} is also pre-trained by image-text pairs but learns to predict the encoded image tokens based on the given text tokens. Conditioned on the input text, DALL-E~\cite{pmlr-v139-ramesh21a} could leverage the \textbf{vision-generative knowledge} to create high-quality synthetic images in a zero-shot manner. Pre-trained by large-scale language corpus, GPT-3~\cite{brown2020language} takes a few hand-written templates as input, and autoregressively generates human-like texts, which contain rich \textbf{language-generative knowledge}. Therefore, the four models have distinctive pre-training goals and can provide complementary knowledge to assist the few-shot visual recognition.

In detail, we cascade them by three steps.: \textbf{1) Prompt}. We adopt GPT-3~\cite{brown2020language} to produce textual prompts for CLIP based on a few hand-written templates. These prompts with richer language knowledge are fed into CLIP's textual encoder. \textbf{2) Generate}. We adopt DALL-E~\cite{pmlr-v139-ramesh21a} to generate additional training images for different categories based on the domain-specific texts, which enlarges the few-shot training data, but costs no extra manpower for collection and annotation. \textbf{3) Cache}. We utilize a cache model to adaptively incorporate the predictions from both CLIP~\cite{radford2021learning} and DINO~\cite{Caron_2021_ICCV}. Referring to Tip-Adapter~\cite{zhang2021tip}, we build the cache model with two kinds of keys respectively for the two pre-trained models. Regarding zero-shot CLIP as the distribution baseline, we adaptively ensemble the predictions of two cached keys as the final output. By only fine-tuning the lightweight cache model via expanded training data, CaFo can learn to fuse diverse prior knowledge and leverage their complementary characteristics for better few-shot visual recognition.

Our main contributions are summarized as follows:
\begin{itemize}
    \item We propose CaFo to incorporate the prior knowledge learned from various pre-training paradigms for better few-shot learning. 
    \item By collaborating CLIP, DINO, GPT-3 and DALL-E, CaFo utilizes more semantic prompts, enriches the limited few-shot training data, and adaptively ensembles diverse predictions via the cache model.
    \item We conduct thorough experiments on 11 datasets for few-shot classification, where CaFo achieves \textit{state-of-the-art} without using extra annotated data.
\end{itemize}

\section{Related Work}

\paragraph{Pre-training of Vision Models.}
With the breakthroughs in deep learning models~\cite{He_2016_CVPR,2010.11929,Liu_2021_ICCV}, most modern vision models are based on the paradigm of pre-training on ImageNet~\cite{deng2009imagenet} and fine-tuning on downstream tasks~\cite{He_2019_ICCV}. Pre-trained models have shown promising adaptability for various downstream tasks, such as object detection~\cite{Lin_2017_ICCV}, semantic segmentation~\cite{7913730}, and 3D recognition~\cite{m2ae,i2p,joint}. 
To improve the representation capability by overcoming the constraints of annotation, self-supervised pre-training has attracted wide attention using large-scale unlabeled datasets ~\cite{9086055}. Self-supervised learning is initialized by pretext tasks, such as image restoration from corruption~\cite{10.1145/1390156.1390294,Pathak_2016_CVPR,He_2022_CVPR}, pseudo labels~\cite{Doersch_2015_ICCV,10.1007/978-3-319-46466-4_5} and clustering~\cite{Caron_2018_ECCV}. Recently, contrast learning, which learns representations by contrasting positive pairs against negative pairs, has gotten well studied for diverse visual representation learning~\cite{He_2020_CVPR,1807.03748,2003.04297,pmlr-v119-chen20j,NEURIPS2020_f3ada80d,Caron_2021_ICCV}. Besides, language-supervised visual pre-training emerges as a novel paradigm closer to natural visual understanding~\cite{Miech_2019_ICCV,1504.00032,sharma-etal-2018-conceptual,Antol_2015_ICCV}, among which CLIP~\cite{radford2021learning} obtains powerful zero-shot transferability by contrastive pre-training on image-text pairs from the Internet. In addition, vision-language pre-training can also promote the zero-shot image generation from text. Open generative models, such as DALL-E~\cite{pmlr-v139-ramesh21a} and CogView~\cite{NEURIPS2021_a4d92e2c} pre-trained on large-scale image-text pairs are able to generate images with diverse contents by given texts. In this paper, CaFo cascade three visual pre-training models, CLIP, DINO, and DALL-E, which contributes to better few-shot learning capacity.


\paragraph{Language-assisted Vision Models.}
As different form of data, linguistic knowledge normally contains complementary knowledge to images. 
For vision-language models, several works\cite{brown2020language, gao2020making, radford2021learning} have showed the format of prompts would highly affect the accuracy on vision tasks. Thus, prompt engineering is worth putting in great effort. Some efforts\cite{rao2022denseclip, huang2022unsupervised,zhou2021learning,zhang2021vt} utilize learnable textual inputs and optimize them during training. Other works\cite{pratt2022does, menon2022visual} propose to leverage linguistic knowledge pre-trained from large language models to generate prompts for each visual category, which enhances vision-language models without any additional training or labeling. Our CaFo refers to CuPL~\cite{pratt2022does} to produce semantic-rich texts to prompt CLIP for better text-image alignment.

\paragraph{Few-shot Learning.}
Few-shot learning highly relies on the transferability of the trained neural networks. 
From the perspective of distance measurement, some metric learning methods learn a metric space by computing the distances from the instances to novel categories~\cite{snell2017prototypical,Sung_2018_CVPR,NIPS2016_90e13578}. Also, meta-learning is proposed to improve the few-shot adaptation ability of the models by finding a set of initialized parameters that can rapidly adapt to novel domains~\cite{pmlr-v70-finn17a,Jamal_2019_CVPR,Chen_2021_ICCV, li2021beyond}. 
More recently, with the vision-language pre-training model CLIP~\cite{radford2021learning} exhibiting strong zero-shot adaptation performance, several efforts have started to find efficient strategies to adapt it to downstream few-shot datasets. CoOp~\cite{coop} is proposed as a prompt tuning adaptation method by optimizing a set of learnable prompt tokens. Subsequently, to inject textual branch with visual signals, CoCoOp~\cite{Zhou_2022_CVPR} and VT-CLIP~\cite{zhang2021vt} propose to train a intermediate network to generate image tokens as conditional inputs for the textual vectors. Referring to adapters~\cite{houlsby2019parameter} in natural language processing, CLIP-Adapter~\cite{2110.04544} is introduced to fine-tune CLIP by applying lightweight residual-style adapters. Tip-Adapter~\cite{zhang2021tip} is then proposed as a training-free adaption method with a constructed key-value cache model. It can also be regarded as a better initialization of CLIP-Adapter with much faster convergence when fine-tuning. CALIP~\cite{guo2022calip} proposes a parameter-free attention to enhance CLIP in a zero-shot manner, and its parametric solution further attains higher few-shot accuracy. SuS-X~\cite{udandarao2022sus} constructs a dynamic support set and extends Tip-Adapter by leveraging image-text distances. Besides, many follow-up works~\cite{zhang2022pointclip,zhu2022pointclip,lin2022frozen,zhang2022collaboration,zhang2022can,huang2022unsupervised,udandarao2022sus} have also been proposed for further adapting CLIP to various vision tasks. Different from all existing methods, we integrate other powerful pre-training paradigms with CLIP and collaborate them with customized pipelines.


\section{Cascade of Foundation Models}

In this section, we first briefly revisit four types of pre-training paradigms in CaFo. Then, we specifically introduce how we cascade them by `Prompt, Generate, then Cache'.

\subsection{Different Pre-training Paradigms}
\label{s3-1}

\paragraph{Contrastive Vision-Language Pre-training.}
The series~\cite{pmlr-v119-chen20j} of contrastive learning between vision and language learn to map the two modalities into the same embedding space via a contrastive loss. Driven by web-scale datasets, e.g., 400 million for CLIP~\cite{radford2021learning} and 1.8 billion for ALIGN~\cite{ALIGN}, the basic pre-training target is to minimize the embedding distances of images and their textual descriptions, while maximize those unpaired ones. By the cross-modal alignment, we can discriminate images of different categorizes by the texts with different semantics. We denote such learned prior as language-contrastive knowledge and adopt CLIP as the representative model for such pre-training method.

\vspace{-12pt}
\paragraph{Contrastive Vision Pre-training.}
As the traditional self-supervised learning methods, vision-contrastive models~\cite{pmlr-v119-chen20j} focus on the discrimination between different images. Normally, the positive pairs to be drawn close are two transformations of the same image, while the optimization of negative pairs~\cite{grill2020bootstrap} is optional, which can be replaced by a momentum encoder~\cite{He_2020_CVPR} or cluster assignments~\cite{caron2020unsupervised}. Recent works reveal that we can learn self-supervised features without negative pairs between images~\cite{NEURIPS2020_f3ada80d,Caron_2021_ICCV}.
Given the strong linear classification capacity, the pre-trained DINO~\cite{Caron_2021_ICCV} is adopted here to provide vision-contrastive knowledge for collaboration.

\vspace{-8pt}
\paragraph{Generative Language Pre-training.}
With 175 billion parameters, the large-scale pre-trained GPT-3\cite{brown2020language} is powerful to produce human-like texts with diverse contents and incredible quality. Taking as input a few designed language commands, GPT-3 is able to output prompts with rich linguistic semantics for vision-language models. 
CLIP utilizes handcrafted templates as prompts, e.g., ``a photo of a [CLASS]'', which however lacks sufficient textual semantics to align with input images. We thus leverage GPT-3 to produce CLIP's prompts to better align with visual information from images.

\vspace{-8pt}
\paragraph{Generative Vision-Language Pre-training.}
Learned from millions of image-caption pairs, the DALL-E series can generate language-conditioned images in a zero-shot manner. They are pre-trained to autoregressively predict the encoded image tokens from the textual tokens of the captions. With such language-generative knowledge, the pre-trained DALL-E can be viewed as a free lunch to enlarge the training data without any manpower. Considering publicity, we select DALL-E-mini~\cite{Mini_Dalle} as the representative among DALL-E models.

\begin{figure}[t]
\centering
\includegraphics[width=\linewidth]{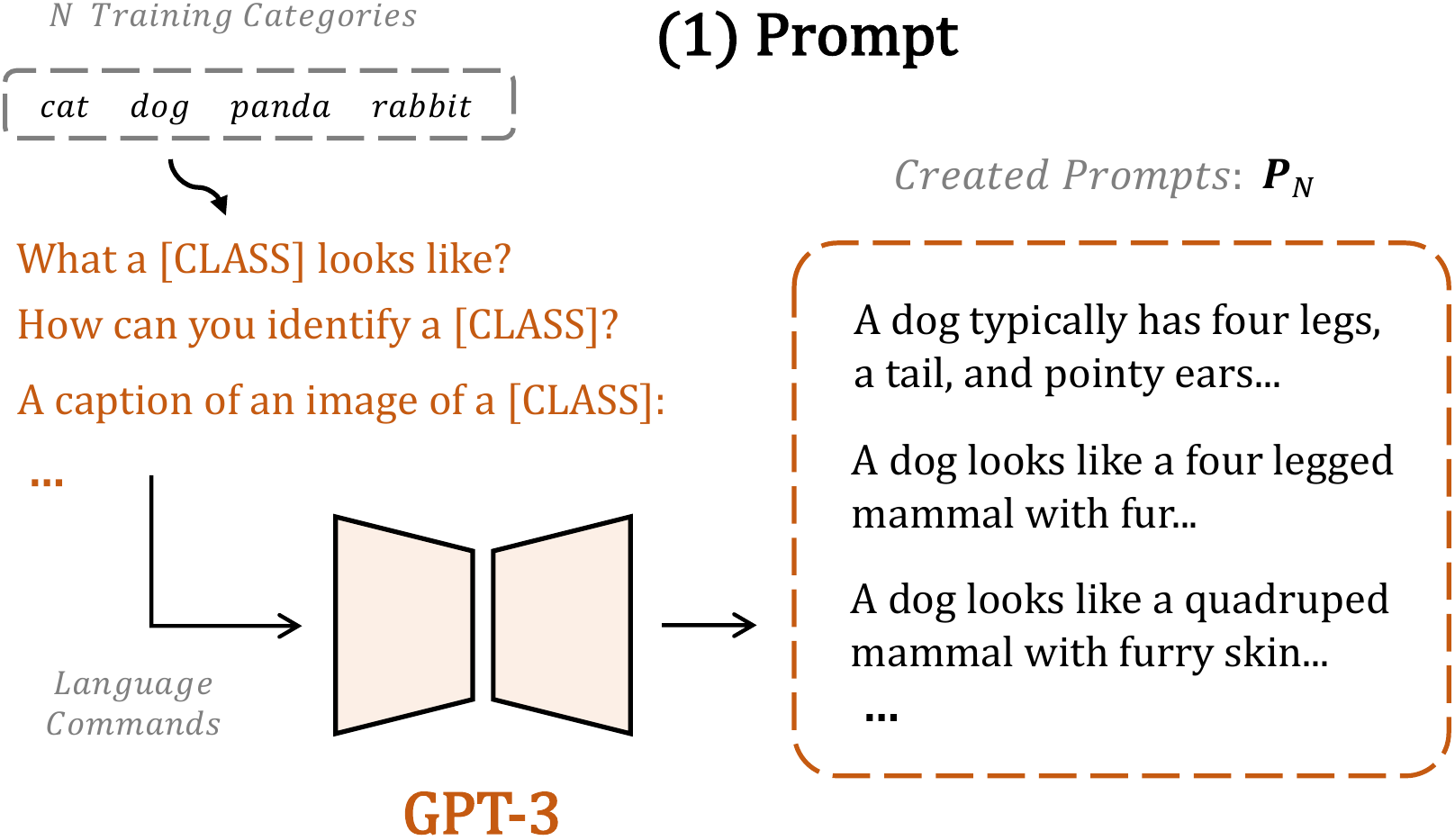} 
\caption{\textbf{Prompt with GPT-3~\cite{brown2020language}}. As the first step in CaFo, we utilize the pre-trained GPT-3 to produce prompts with rich linguistic semantics for CLIP's textual encoder.}
\label{fig:GPTframework}
\vspace{-8pt}
\end{figure}

\begin{figure*}[t]
\centering
\includegraphics[width=0.95\textwidth]{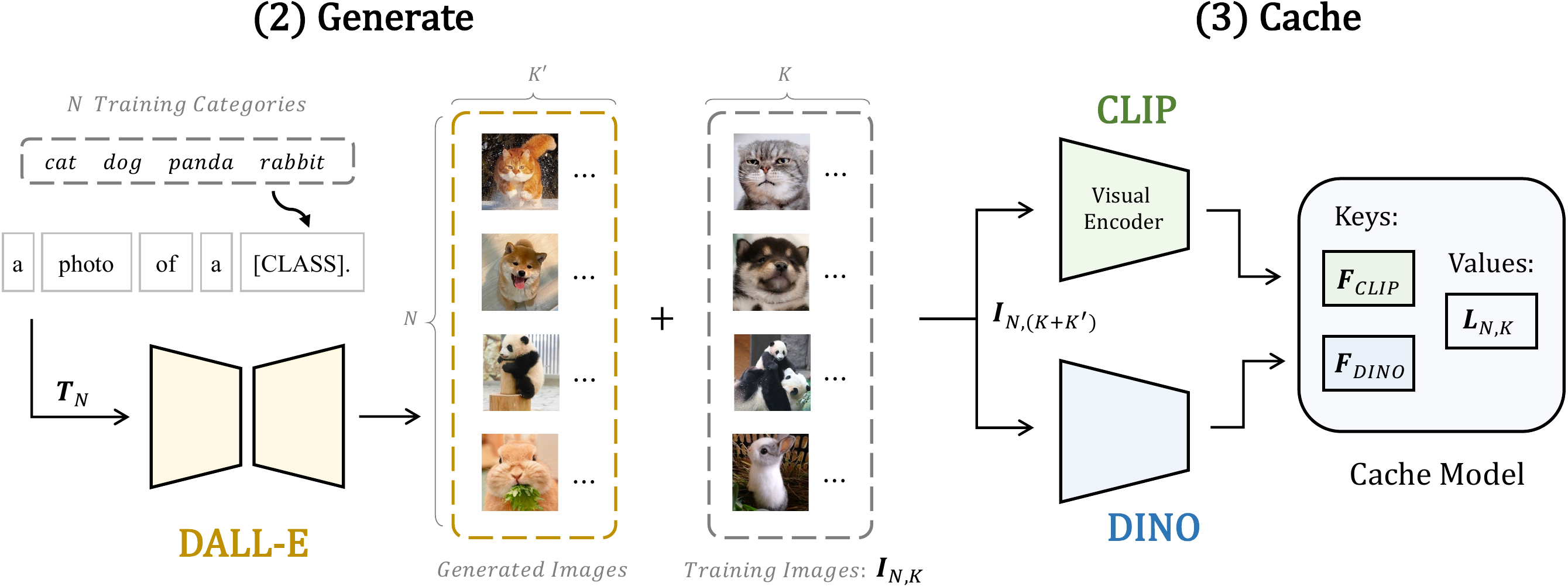} 
\vspace{0.1cm}
\caption{\textbf{Generate via DALL-E~\cite{pmlr-v139-ramesh21a}, then Cache by CLIP~\cite{radford2021learning} and DINO~\cite{Caron_2021_ICCV}.}
We adopt DALL-E to generate synthetic images to expand the limited few-shot training samples. Then, we construct the cache model with two kinds of keys to adaptively fuse the knowledge from CLIP and DINO.}
\label{fig:Framework1}
\end{figure*}

\subsection{Prompt, Generate, then Cache}

To cascade different pre-training paradigms, we introduce CaFo with a pipeline of `Prompt, Generate, then Cache', which respectively unleashes the powers of different self-supervised knowledge.

\paragraph{Prompt with GPT-3.}
Under the $N$-way $K$-shot settings, we have the few-shot training images $I_{N,K}$ with labels $L_{N,K}$ that contain $K$ samples for each $N$ categories. As shown in Figure~\ref{fig:GPTframework}, for $N$ categories, we adopt a unified series of templates as the language command for GPT-3~\cite{brown2020language}, e.g., ``What a [CLASS] looks like?'', ``How can you identify a [CLASS]?'', and ``A caption of an image of a [CLASS]:''. We denote the created prompts for $N$ categories as $P_N$, formulated as
\begin{align}
    P_{N}= \text{GPT-3}(\text{\textcolor{gray}{Commands}}).
   \label{eq:gpt3}
\end{align}
Then, we adopt $P_{N}$ as the input of CLIP's textual encoder.
Further, for some downstream data with specialized categories, we can customize the language commands for producing prompts with more domain-specific semantics. For example, in OxfordPets~\cite{parkhi2012cats} dataset of pet images, we adopt the input of GPT-3 as ``This is a pet bulldog, it has thin neck, short face, floppy ears. It's coat is short, straight, and in brindle color. This is a pet [CLASS],''. Based on that, GPT-3 continues to describe more details of the [CLASS] pet.


\paragraph{Generate via DALL-E}
Via the zero-shot DALL-E~\cite{Mini_Dalle}, we generate synthesis images to enrich our limited training images $I_{N,K}$, as shown in Figure~\ref{fig:Framework1} (1). For different categories, we adopt a simple template, e.g., ``a photo of a [CLASS].''. After the generation, we utilize CLIP to filter the top-$K'$ best-quality images as the newly-expanded training samples for each category. Then, we obtain the $N$-category ($K+K'$)-sample training images, formulated as
\begin{align}
    I_{N,(K+K')}= \{\text{DALL-E}(T_{N}),\ \ I_{N,K}\},
   \label{eq:dalle},
\end{align}
where $T_N$ denotes the $N$-category textual inputs.
We keep $K'$ comparable with $K$ to ensure the synthesis quality and also preserve the low-data regimes. By the pre-trained language-generative knowledge, the data expansion is totally zero-shot, which requires no manpower to collect or annotate the data, and alleviates the data deficiency issue inherently for few-shot learning.

\begin{figure*}[t]
\centering
\includegraphics[width=0.95\textwidth]{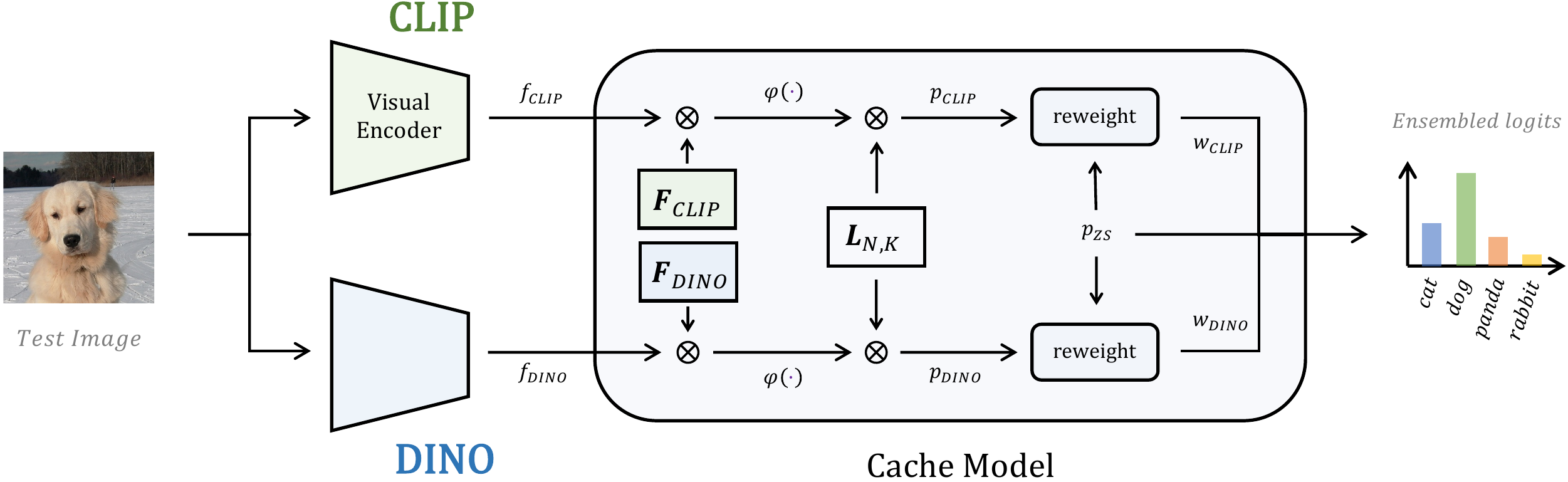} 
\caption{\textbf{Adaptive Inference with Cache Model.} We regard the test image as a query and retrieves CLIP and DINO's knowledge from the corresponding two keys in the cache model. Then, we calculate the distribution similarities between different classification logits for adaptive ensemble.}
\label{fig:Framework2}
\vspace{-8pt}
\end{figure*}


\paragraph{Cache by CLIP and DINO.} We construct a key-value cache model for adaptive knowledge ensemble. Different from Tip-Adapter~\cite{zhang2021tip} only adapting CLIP, our cache model contains the pre-learned knowledge from both CLIP and DINO by caching two kinds of keys. Specifically in Figure~\ref{fig:Framework2} (2), we first utilize CLIP and DINO to independently extract visual features of the few-shot training images, formulated as
\begin{align}
    F_{\text{CLIP}} &= \text{CLIP$_{vis}$}(I_{N,(K+K')});\\
    F_{\text{DINO}} &= \text{DINO}(I_{N,(K+K')}),
   \label{eq:extract}
\end{align}
where CLIP$_{vis}$ denotes the CLIP's visual encoder and $F_{\text{CLIP}}, F_{\text{DINO}}\in \mathbb{R}^{N(K+K')\times C}$. Besides the two keys, we convert the few-shot training labels into one-hot encodings $L_{onehot}\in \mathbb{R}^{N(K+K')\times N}$, and regard them as the same values for both keys. During training, we follow Tip-Adapter that only enables the cached keys in the adapter to be learnable and keeps the pre-trained models frozen.

\subsection{Adaptive Inference}

For a test image in Figure~\ref{fig:Framework2}, we first extract its two visual features $f_{\text{CLIP}},f_{\text{DINO}}\in \mathbb{R}^{1\times C}$ and regard them as queries to retrieve diverse knowledge from the cache model. Then, we could acquire three predicted classification logits $p_{\text{ZS}},p_{\text{CLIP}},p_{\text{DINO}}\in \mathbb{R}^{1\times N}$, which are respectively from CLIP's zero-shot alignment and the two keys of cache model. We formulate them as
\begin{align}
    p_{\text{ZS}} &= f_{\text{CLIP}}\ \text{CLIP$_{tex}$}(P_N)^T;\\
    p_{\text{CLIP}} &= \varphi(f_{\text{CLIP}}F_{\text{CLIP}}^T)\ L_{onehot};\\
    p_{\text{DINO}} &= \varphi(f_{\text{DINO}}F_{\text{DINO}}^T)\ L_{onehot},
   \label{eq:logits}
\end{align}
where CLIP$_{tex}$ represents CLIP's textual encoder, $P_N$ denotes GPT-3's created prompts, and $f_{\text{CLIP}}F_{\text{CLIP}}^T$ denotes the query-key affinity matrix of the CLIP's keys, analogous to DINO's. $\varphi(x) = \exp(-\beta\cdot(1-x))$ serves as a non-linear modulator to control the sharpness of affinity matrix.

As the language-contrastive $p_{\text{ZS}}$ is pre-trained by 400 million data and can perform strong zero-shot transfer ability, we regard $p_{\text{ZS}}$ as the prediction baseline and calculate the weights of $p_{\text{CLIP}},p_{\text{DINO}}$ for ensemble based on their distribution similarity with $p_{\text{ZS}}$. By this, we can suppress some obviously false category possibilities in $p_\text{CLIP},p_\text{DINO}$ and also amplify the moderately correct ones during ensemble. Firstly, we respectively normalize the scales of three classification logits into -1$\sim$1 by their each mean and standard deviation. We then calculate the distribution similarities as the ensemble weights for the two logits of the cache as
\begin{align}
    w_{\text{CLIP}} = p_{\text{CLIP}}\ p_{\text{ZS}}^T;\ \ w_{\text{DINO}} = p_{\text{DINO}}\ p_{\text{ZS}}^T.
   \label{eq:sim}
\end{align}
Finally, we adopt the softmax function to normalize the weights and obtain the final ensemble logits as
\begin{align}
    p_{en} = p_{\text{ZS}} + \sum_{i}p_i\cdot\text{softmax($w_i$)},
   \label{eq:sim}
\end{align}
where $i\in \{\text{CLIP}, \text{DINO}\}$. By such similarity-based ensemble, $p_{en}$ can adaptively fuse the prior knowledge learned by CLIP and DINO's pre-training and achieve stronger few-shot image classification.

\section{Experiments}
\subsection{Settings}
\paragraph{Datasets.} We conduct few-shot experiments on 11 publicly available datasets: ImageNet~\cite{deng2009imagenet}, StandfordCars~\cite{krause20133d}, UCF101~\cite{soomro2012ucf101}, Caltech101~\cite{fei2004learning}, Flowers102~\cite{nilsback2008automated}, SUN397~\cite{xiao2010sun}, DTD~\cite{cimpoi2014describing}, EuroSAT~\cite{helber2019eurosat}, FGVCAircraft~\cite{maji2013fine}, OxfordPets~\cite{parkhi2012cats}, and Food101~\cite{bossard2014food}. We follow Tip-Adapter \cite{zhang2021tip} to train CaFo with 1, 2, 4, 8, 16 shots and test on the full test set. \textit{As we adopt DALL-E to generate training images in a zero-shot manner, we can train CaFo only by the generated images and report its zero-shot performance without few-shot training set.}

\vspace{-15pt}
\paragraph{Implementation.}
Our CaFo integrates the knowledge from pre-trained CLIP~\cite{radford2021learning}, DINO~\cite{Caron_2021_ICCV}, DALL-E~\cite{pmlr-v139-ramesh21a}, and GPT-3~\cite{brown2020language}. For CLIP, we utilize ResNet-50~\cite{He_2016_CVPR} as the visual encoder and its aligned transformer as the textual encoder. To align with the visual representation from CLIP, we also adopt DINO pre-trained upon ResNet-50. For DALL-E, we adopt different domain-specific textual templates as the input for different datasets, which correspond to the original textual prompts for CLIP's textual encoder. For GPT-3, we adopt five simple templates as the language commands shared by different categories. Each command outputs ten prompts, which obtains fifty prompts in total. For each category, we simply ensemble the features of different prompts following CuPL~\cite{brown2020language}. During training, we only set the two kinds of keys in cache model to be learnable and utilize the data augmentation following Tip-Adapter-F. We train CaFo using batch size 64 only for 20 epochs, and adopt AdamW optimizer with the initial learning rate 0.0001 with a cosine scheduler. Note that, we tune the hyperparameters in CaFo by the official validation sets.

\begin{figure}[t]
\centering
\vspace{-0.5cm}
\includegraphics[width=0.75\linewidth]{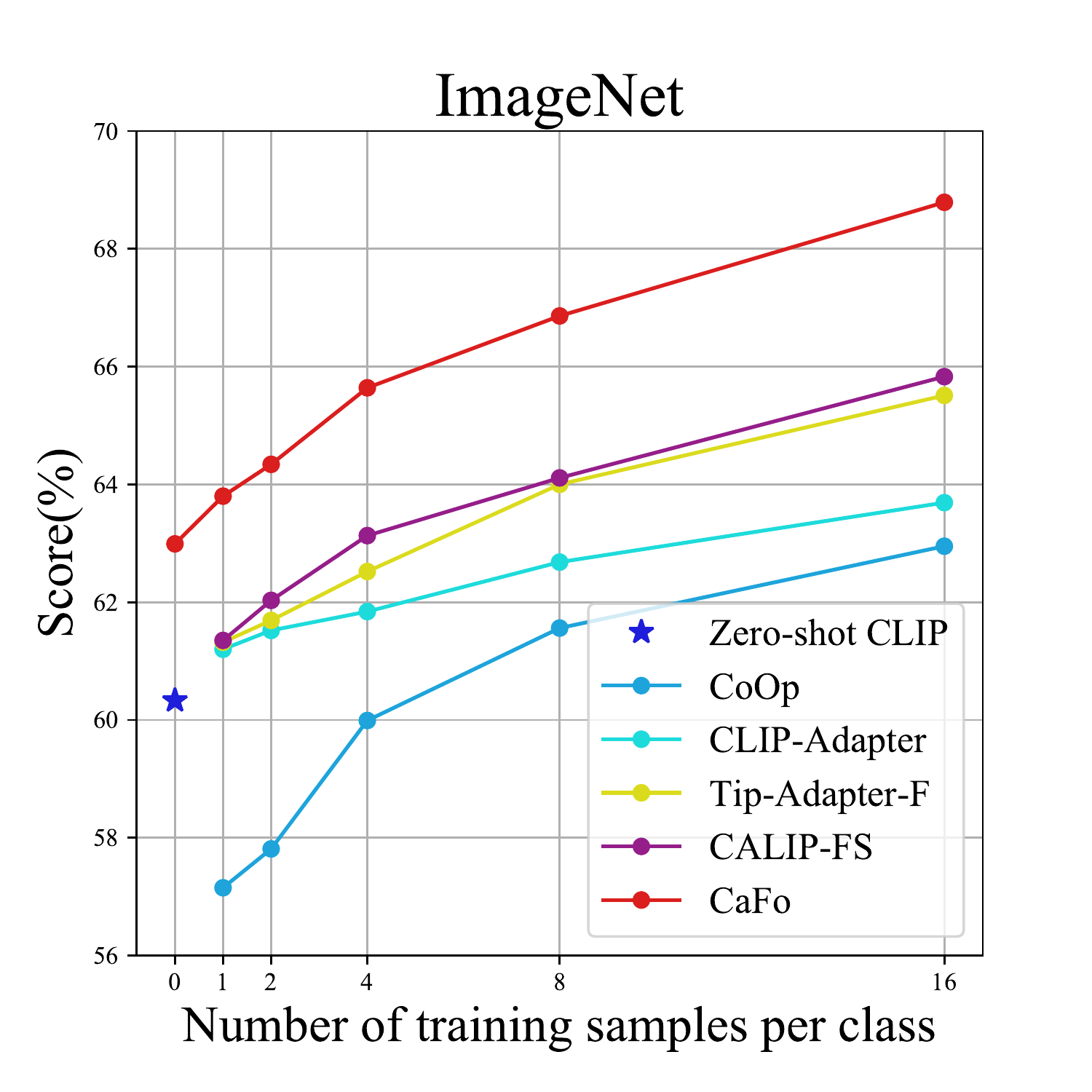}
\caption{\textbf{Performance (\%) Comparison on ImageNet.} We compare CaFo with other methods for different few-shot settings.}
\label{fig:Imagenet Performance comparison}
\end{figure}

\setlength{\tabcolsep}{3pt}
    \begin{table}[t]
    \centering
    \vspace{0.1cm}
    \begin{tabular}{lcccc}
    \toprule
        \multirow{1}*{Models} &Epochs &Time &Accuracy &Gain\\
        \cmidrule(lr){1-1}
        \cmidrule(lr){2-5}
         Zero-shot CLIP & 0 & 0 & 60.33 & -\\
         Zero-shot CALIP & 0 & 0 & 60.57 & -\\
         \cmidrule(lr){1-5} 
         Linear-probe CLIP & - & 13min & 56.13 &-4.20\\
         CoOp &200 &14h\ 40min &62.95 &+2.62\\
         CLIP-Adapter &200 &50min &63.59 &+3.26\\
         Tip-Adapter-F &\textbf{20} &\textbf{5min} &65.51 &+5.18\\
         CALIP-FS &200 &1h &65.81 &+5.48\\\cmidrule(lr){1-5} 
         \textbf{CaFo} &\textbf{20} &10min &\bf68.79 &\bf+8.46\\
         \bottomrule
    \end{tabular}
    \vspace{0.05cm}
    \caption{\textbf{Efficiency Comparison on ImageNet.} We test the training time with a single A100 GPU under 16-shot setting.}
    \label{tab:time comparison}
    \end{table}
    \setlength{\tabcolsep}{1.4pt}
    

\subsection{Performance}

\paragraph{On ImageNet.}
We compare CaFo with other CLIP-based adaption methods on the most representative ImageNet~\cite{deng2009imagenet}: CALIP~\cite{guo2022calip}, Linear-probe CLIP~\cite{radford2021learning}, CoOp~\cite{coop}, CLIP-Adapter~\cite{2110.04544}, Tip-Adapter-F~\cite{zhang2021tip}, and CALIP-FS~\cite{guo2022calip}. All these methods are based on the pre-trained CLIP~\cite{radford2021learning} with ResNet-50 visual encoders. 
As reported in Figure~\ref{fig:Imagenet Performance comparison} and Table~\ref{tab:imagenet performance comparison}, CaFo surpasses all existing methods for different shot settings.
Remarkably, CaFo with 1 shot even outperforms the 8-shot Linear-probe CLIP and CoOp, and CaFo with 8 shots is better than all methods with 16 shots.
For zero-shot learning, CaFo sigiificantly surpasses CLIP and CALIP, demonstrating the importance of DALL-E's generation.
In Table~\ref{tab:time comparison}, we present the efficiency of CaFo concerning training epochs and time. Our CaFo achieves the best performance-efficiency trade-off with 68.79\% accuracy and only 10 minutes training.
\setlength{\tabcolsep}{2pt}
    \begin{table}[t]
    \centering
    \vspace{0.4cm}
    \begin{tabular}{lcccccc}
    \toprule
        \multirow{1}*{Shot} &0 &1 &2 &4 &8 &16\\
        \cmidrule(lr){1-1}
        \cmidrule(lr){2-7}
         Zero-shot CLIP &60.33 &-&-&-&-&-\\
         Zero-shot CALAP &60.57 &-&-&-&-&-\\
         \cmidrule(lr){1-7} 
         Linear-probe CLIP &- &22.17 & 31.90 &41.20 &49.52 &56.13\\
         CoOp &- &57.15 &57.81 &59.99 &61.56 &62.95\\
         CLIP-Adapter &- &61.20 &61.52 &61.84 &62.68 &63.59\\
         VT-CLIP &- &60.53 &61.29 &62.02 &62.81 &63.92\\
         Tip-Adapter-F &- &61.32 &61.69 &62.52 &64.00 &65.51\\
         CALIP-FS &- &61.35 &62.03 &63.13 &64.11 &65.81\\
        \cmidrule(lr){1-7} 
         \textbf{CaFo} &\bf62.99 &\bf63.80 &\bf64.34 &\bf65.64 &\bf66.86 &\bf68.79 \\
         \bottomrule
    \end{tabular}
    \caption{\textbf{Quantative Performance (\%) Comparison on ImageNet.} For zero-shot performance, CaFo is trained with images generated by DALL-E without any few-shot data.}
    \vspace{-0.1cm}
    \label{tab:imagenet performance comparison}
    \end{table}
    \setlength{\tabcolsep}{1.4pt}

\begin{figure*}[t!]
    \centering
    \subfloat{\includegraphics[scale=0.225]{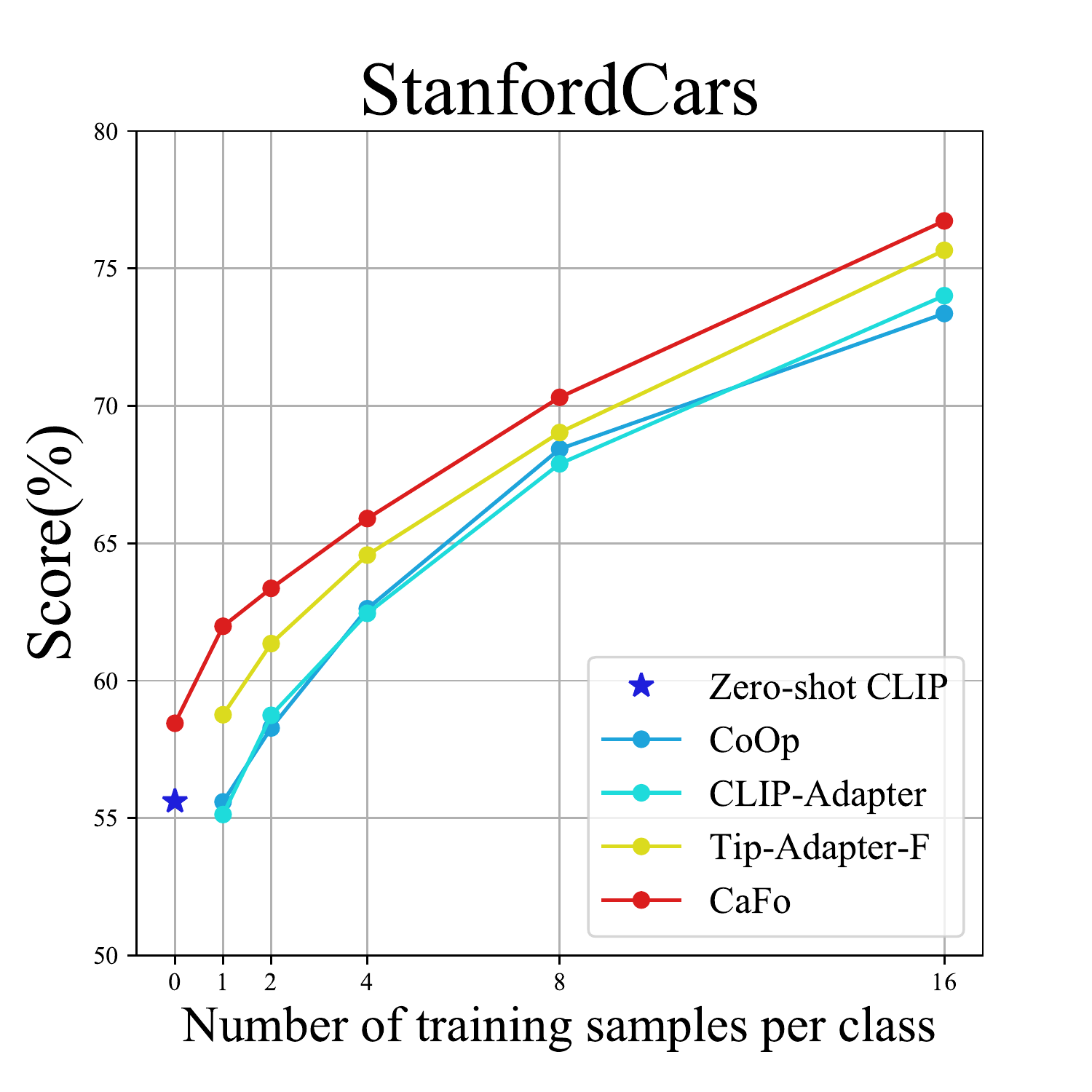}}
    \subfloat{\includegraphics[scale=0.225]{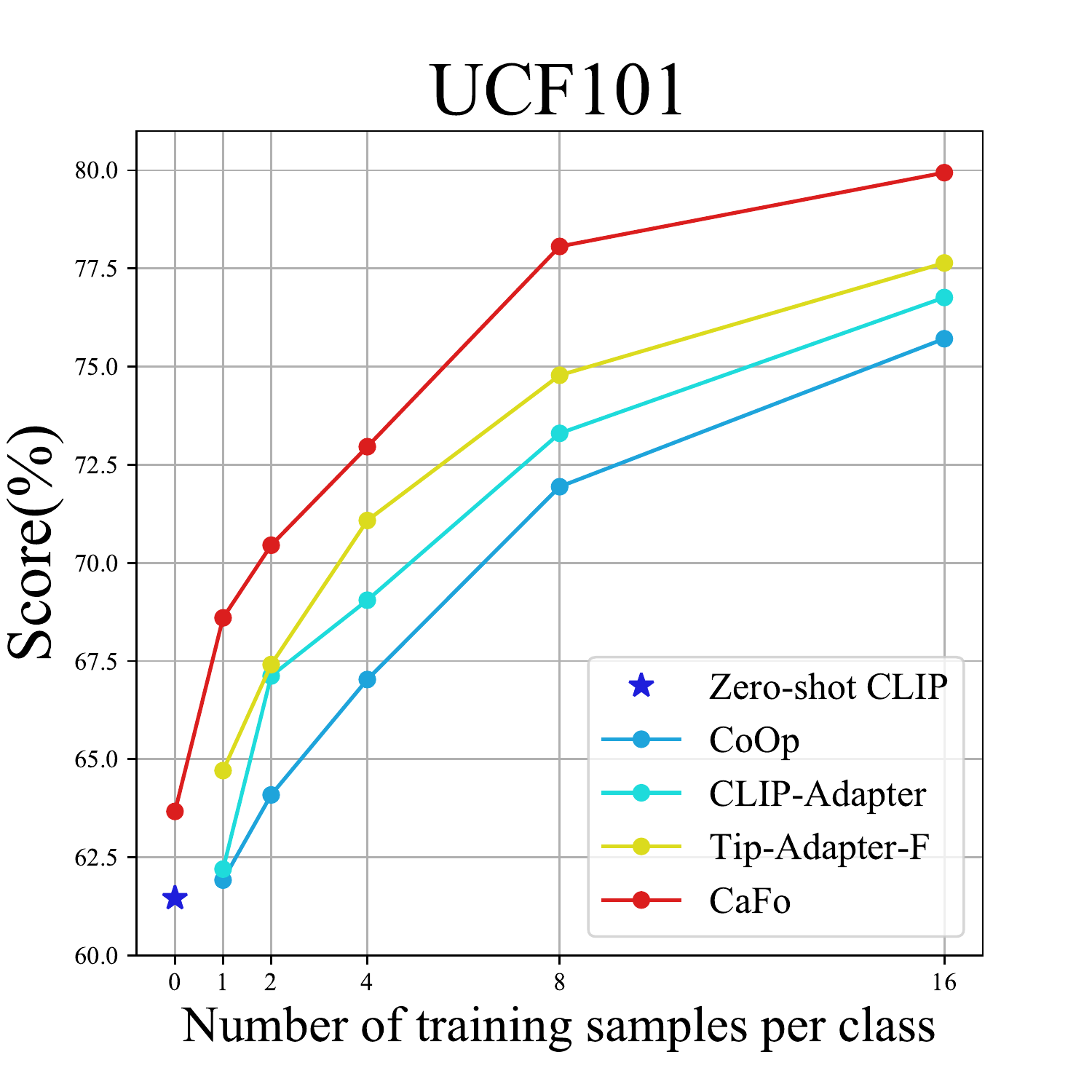}}
    \subfloat{\includegraphics[scale=0.225]{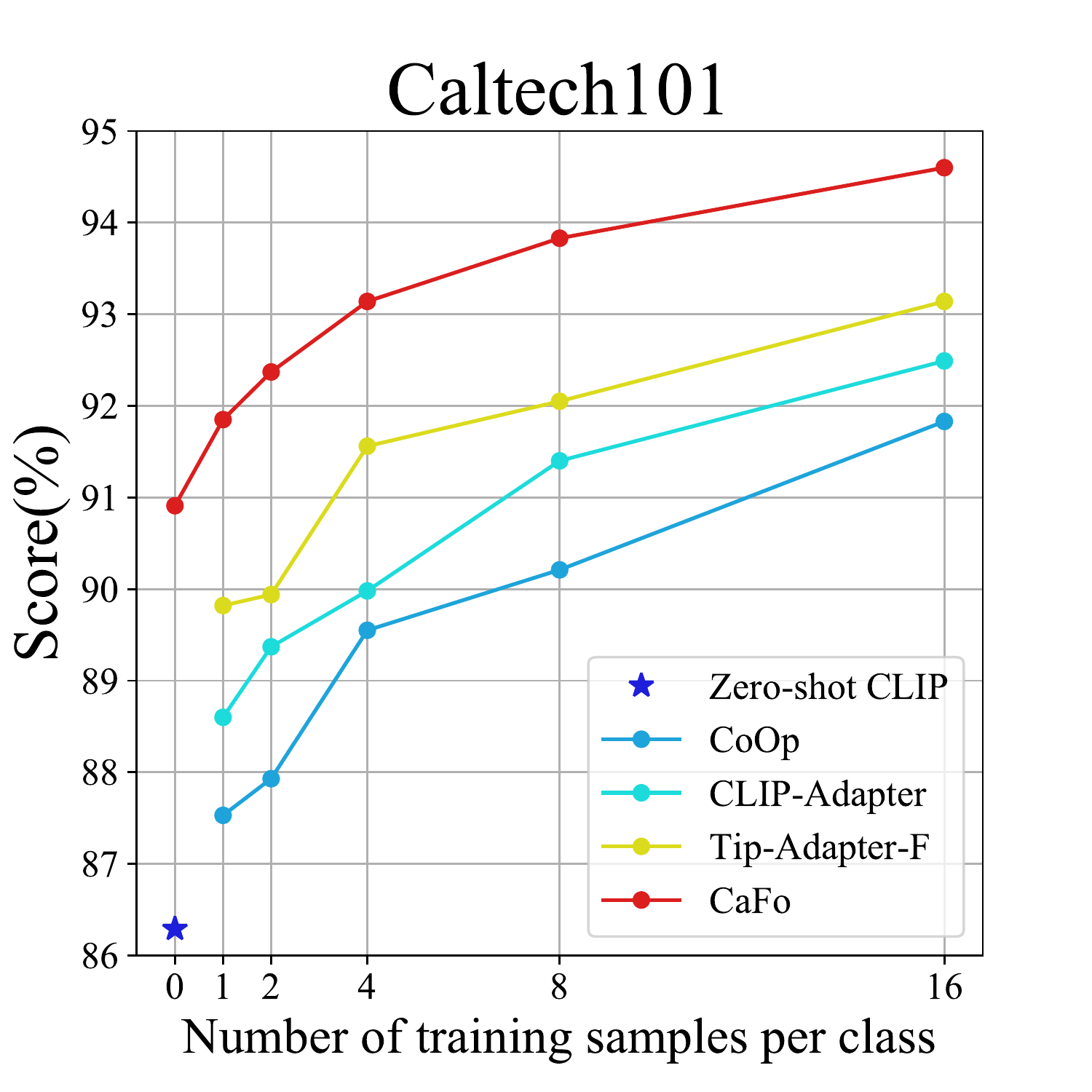}}
    \subfloat{\includegraphics[scale=0.225]{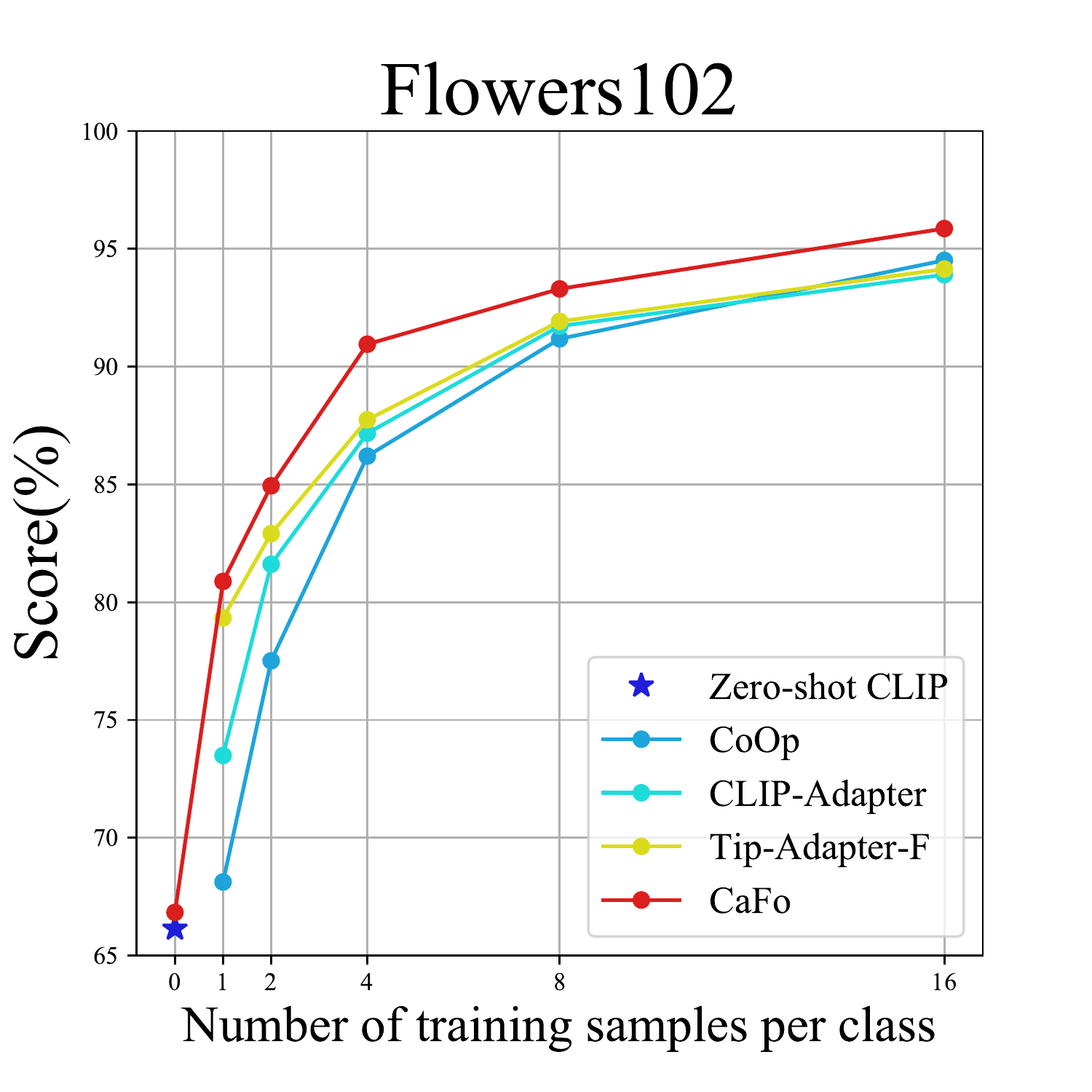}}
    \subfloat{\includegraphics[scale=0.225]{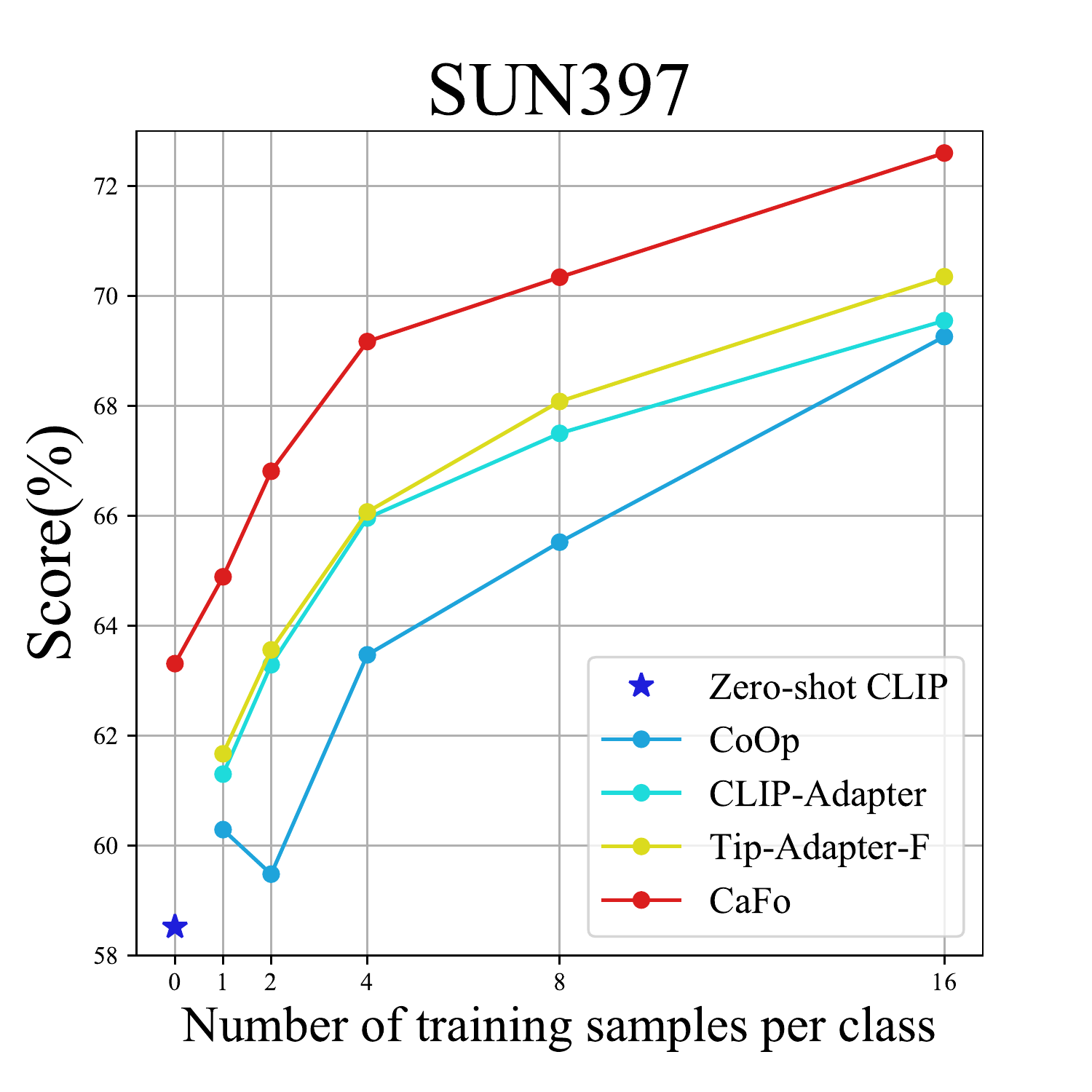}}
    \\
    \subfloat{\includegraphics[scale=0.225]{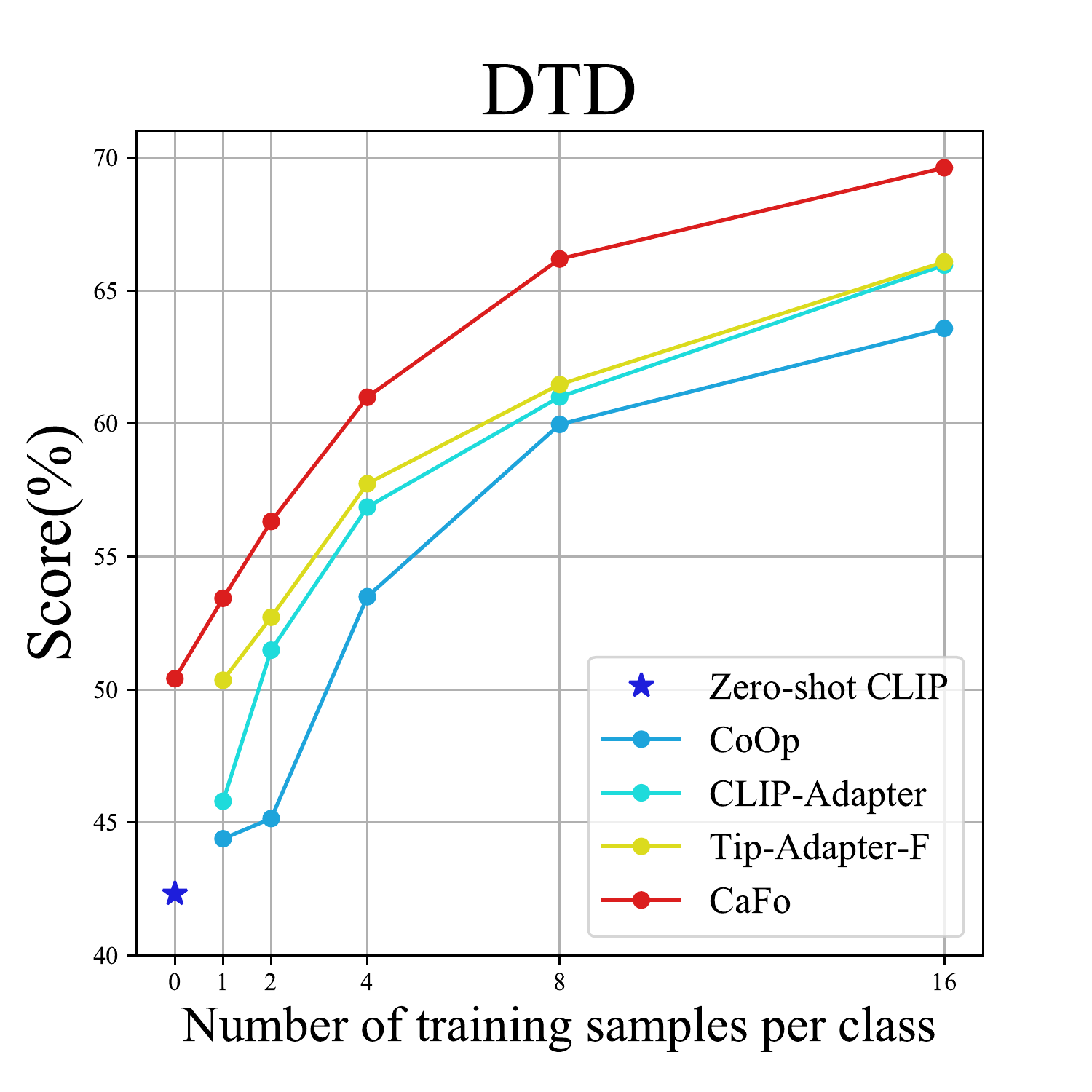}}
    \subfloat{\includegraphics[scale=0.225]{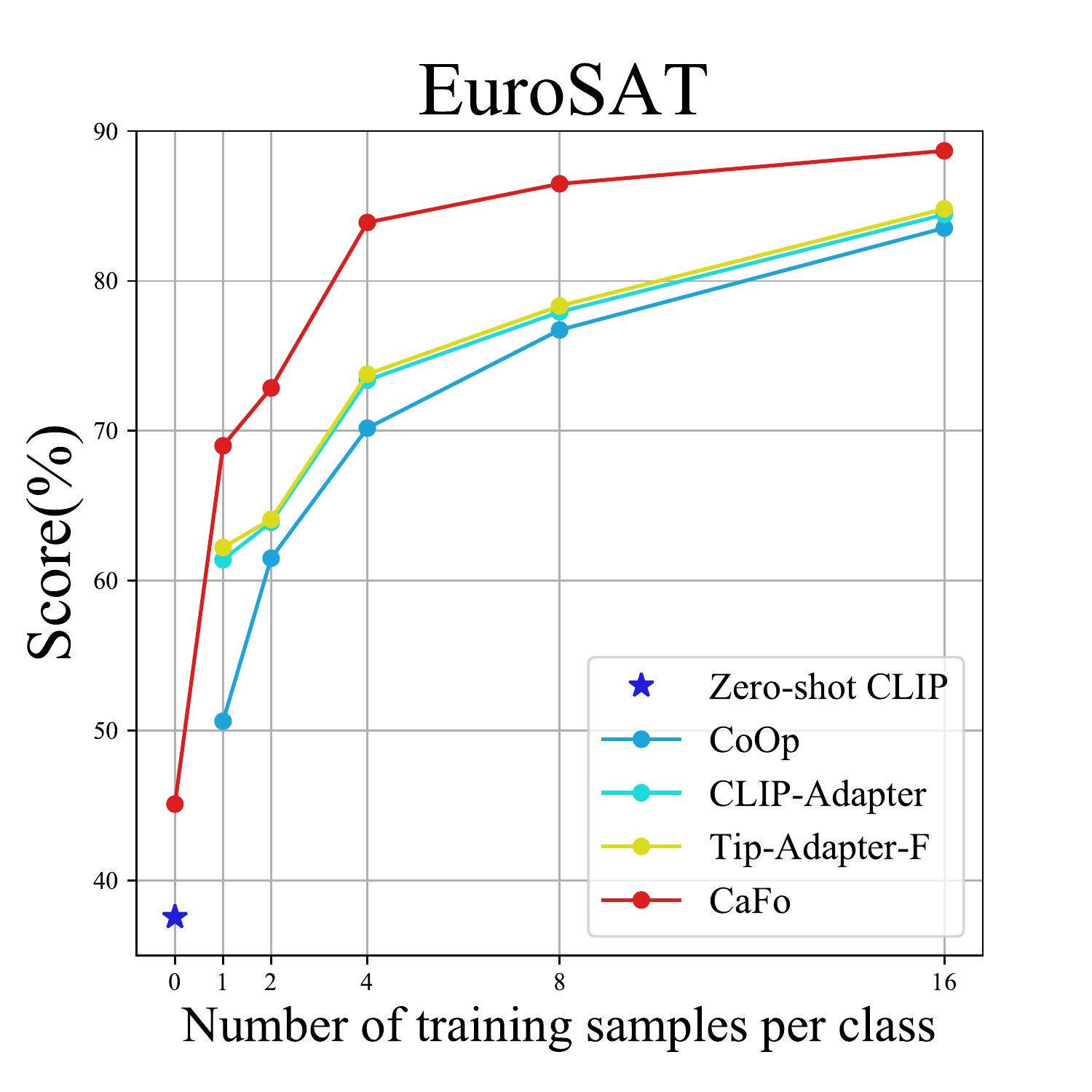}}
    \subfloat{\includegraphics[scale=0.225]{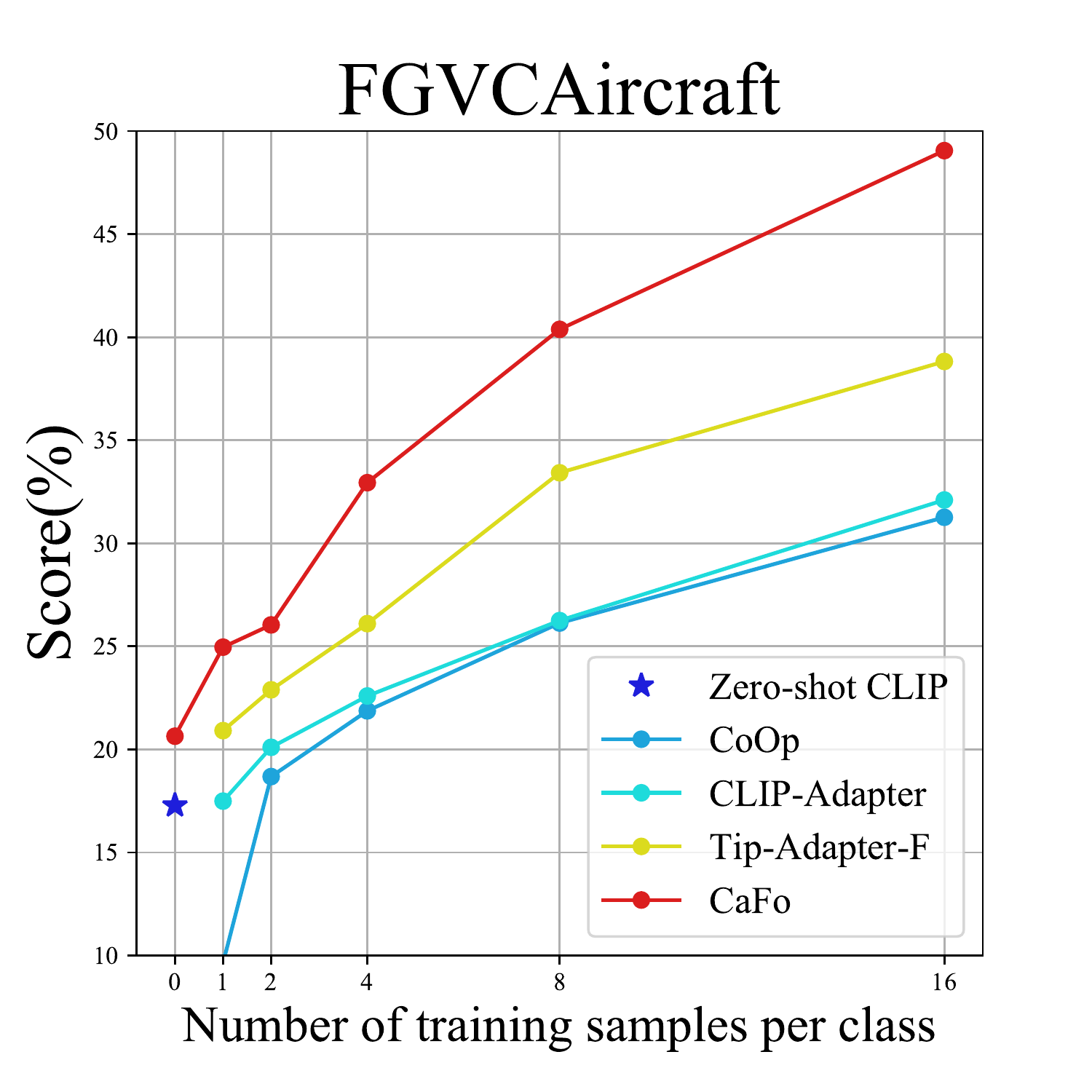}}
    \subfloat{\includegraphics[scale=0.225]{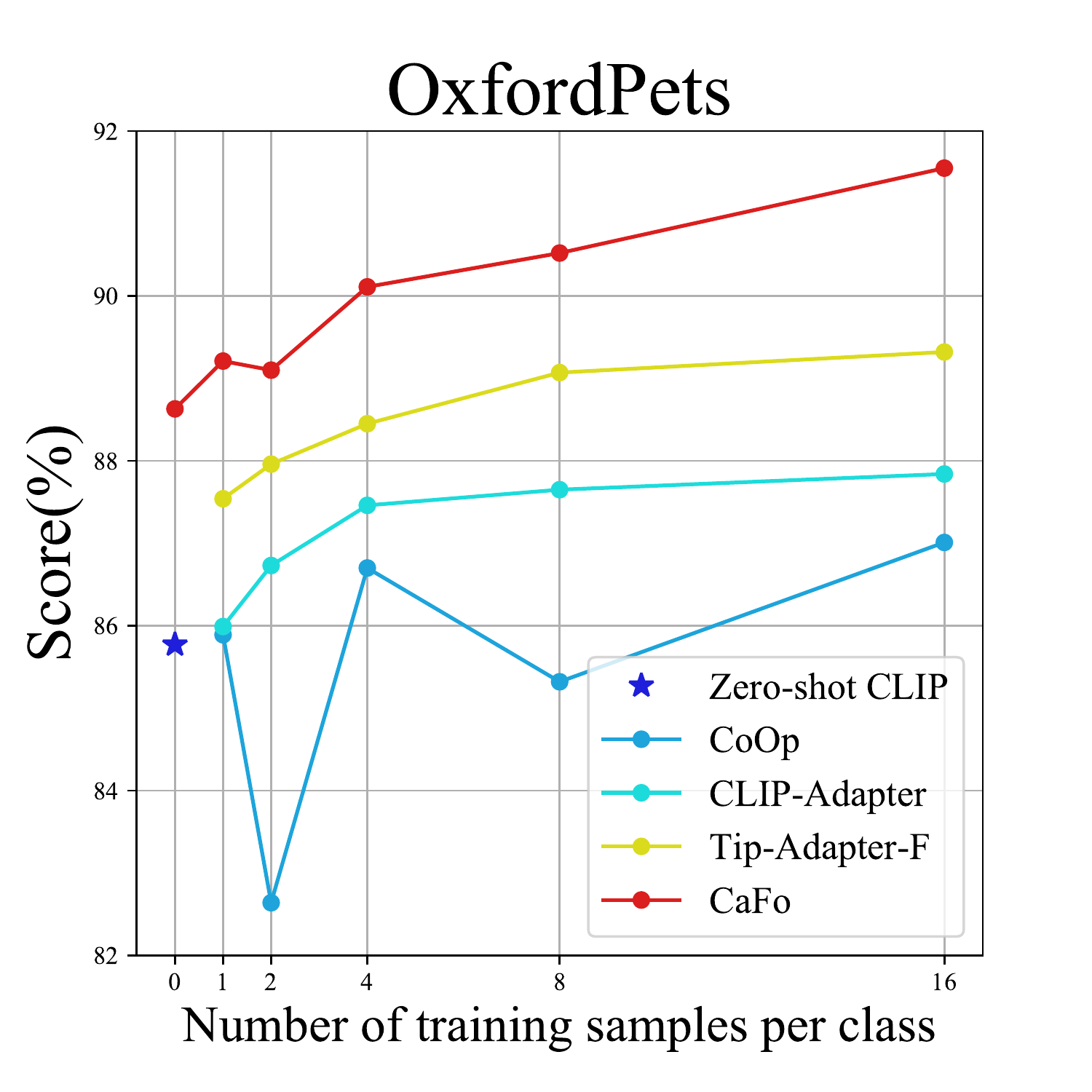}}
    \subfloat{\includegraphics[scale=0.225]{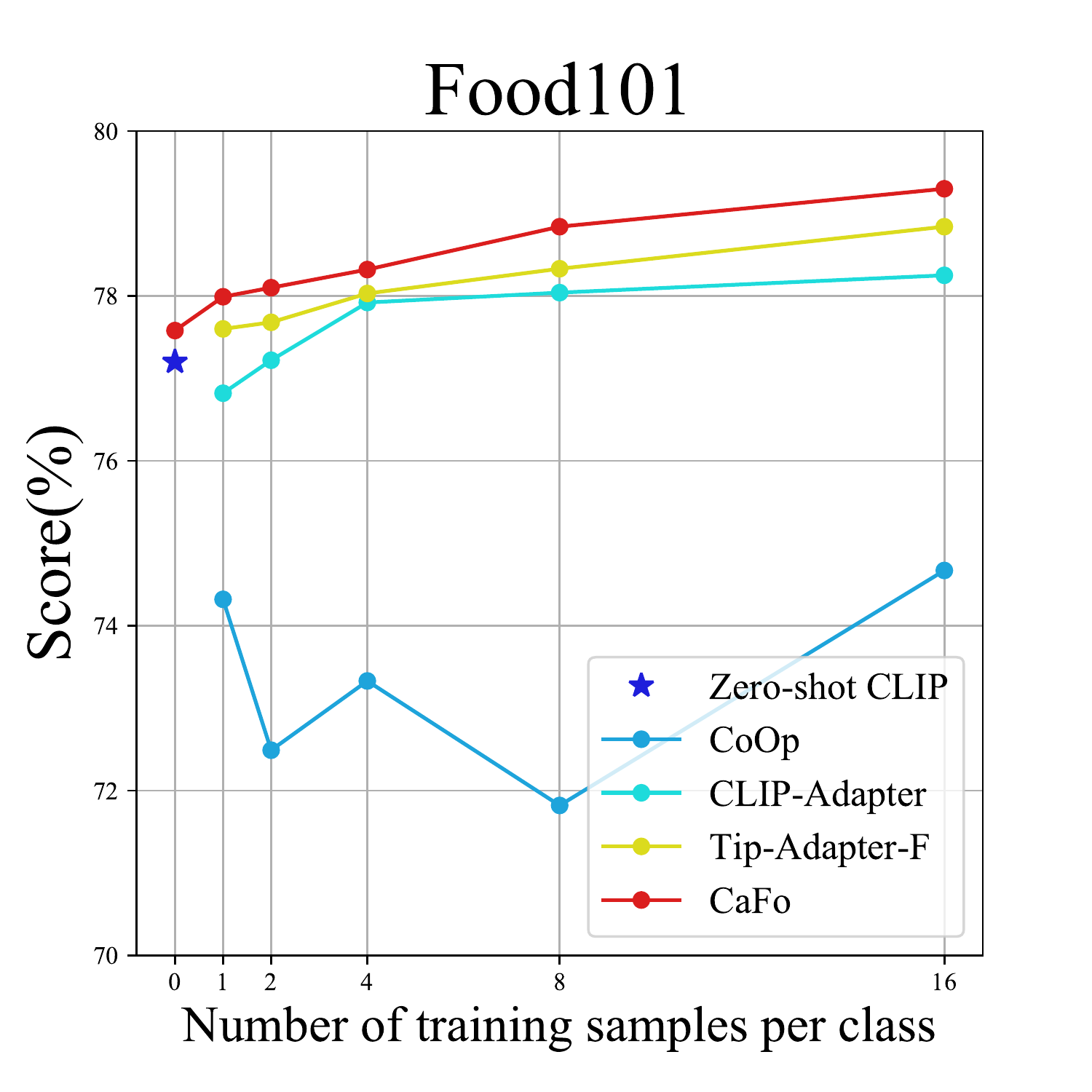}}
    \caption{\textbf{Performance (\%) Comparison on 10 Datasets.} Our method shows \textit{state-of-the-art} performance for all few-shot settings on different datasets, which indicates superior generalization capacity.}
    \label{fig:other_dataset}
\end{figure*}
\vspace{8pt}

\paragraph{On Other Datasets.}
To further assess the robustness in different scenarios, we test CaFo on extra 10 datasets in Figure \ref{fig:other_dataset}. For different semantic domains including real-world scenes, detailed textures, and satellite-captured landscapes, CaFo consistently shows leading performance and indicates excellent robustness via the collaboration of diverse knowledge. Notably, on some datasets, e.g., Caltech101 and OxfordPets, the zero-shot CaFo perform even comparably to other methods with 4 shots, demonstrating the effectiveness of zero-shot DALL-E for few-shot data expansion. 

\setlength{\tabcolsep}{8pt}
\small
\begin{table}[t]
\begin{tabular}{lccc}
\toprule
\multirow{2}{*}{Datasets} & \textbf{Source} &\multicolumn{2}{c}{\textbf{Target}} \\
\cmidrule(lr){2-2} \cmidrule(lr){3-4} 
& ImageNet  & -V2 & -Sketch  \\ \midrule
Zero-shot CLIP  & 60.33  & 53.27 & 35.44\\
Zero-shot CALIP  & 60.57  & 53.70 & 35.61\\
\cmidrule(lr){1-4} 
CoOp & 62.95  & 54.58 & 31.04  \\
CLIP-Adapter &  {63.59}  &  {55.69} &  {35.68} \\
CALIP-FS & 65.81  & 55.98 & 35.37 \\
Tip-Adapter-F & 65.51  & 57.11 & 36.00 \\\cmidrule(lr){1-4} 
\textbf{CaFo}            & \textbf{68.79} & \textbf{57.99} & \textbf{39.43}\\
\bottomrule
\end{tabular}
\caption{\textbf{Distribution Shift (\%) Comparison.} We train the models on ``Source'' dataset and test on ``Target'' datasets.}
\label{tab:domain shifts}
\end{table}
\setlength{\tabcolsep}{1.4pt}
\vspace{-8pt}

\paragraph{Distribution Shift.}
We further evaluate the robustness of CaFo to distribution shift by training on ``Source'' dataset and testing on ``Target'' datasets. In Table~\ref{tab:domain shifts}, we select the ``Source'' as ImageNet and the ``Target'' as ImageNet-V2~\cite{recht2019imagenet} and ImageNet-Sketch~\cite{hendrycks2021natural}. As we can utilize some prior knowledge of the target domain for GPT-3 and DALL-E for prompting and generation, CaFo achieves the best out-of-distribution performance on the two ``Target'' datasets, surpassing the second-best Tip-Adapter-F by +3.28\%, +0.88\%, and +3.43\%, respectively.

\setlength{\tabcolsep}{3pt}
    \begin{table}[t]
    \centering
    \small
    \begin{center}
    \begin{tabular}{ccccccc}
    \toprule\noalign{\smallskip}
    \multicolumn{4}{c}{{Pre-trained Models}} & \multicolumn{3}{c}{{Shot}}\\
    \noalign{\smallskip}
    \cmidrule(lr){1-4}
    \cmidrule(lr){5-7}
    \noalign{\smallskip}
    CLIP &DINO & DALL-E & GPT-3 &1 &4 &16 \\
    \noalign{\smallskip}
    \cmidrule(lr){1-7}
    \noalign{\smallskip}
    \multicolumn{1}{c}{\checkmark} &{} &{} &{} & 61.32  & 62.52 & 65.51 \\ 
    {} & \multicolumn{1}{c}{\checkmark} &{} &{} & 34.14  & 40.47  & 53.27 \\
    \multicolumn{1}{c}{\checkmark} & \multicolumn{1}{c}{\checkmark} &{} &{} & 61.39  &63.96 & 68.08 \\
    \multicolumn{1}{c}{\checkmark} & \multicolumn{1}{c}{\checkmark} & \multicolumn{1}{c}{\checkmark} &{} &63.24  &65.23  &68.42\\
    \multicolumn{1}{c}{\checkmark} & \multicolumn{1}{c}{\checkmark} &  &\multicolumn{1}{c}{\checkmark} &62.32  &64.64  &68.51\\
    \cmidrule(lr){1-7}
    \multicolumn{1}{c}{\checkmark} & \multicolumn{1}{c}{\checkmark} & \multicolumn{1}{c}{\checkmark} &\multicolumn{1}{c}{\checkmark} &\bf63.80  &\bf65.64  &\bf68.79\\
    \bottomrule
    \end{tabular}
    \end{center}
    \vspace{-8pt}
    \caption{\textbf{Ablation Study (\%) of Cascaded Models.} We ablate different pre-trained models on ImageNet with 1, 4, and 16 shots.}
    \label{table:multi pretrain model ablation module}
    \end{table}
    \setlength{\tabcolsep}{1.4pt}

\subsection{Ablation study}

\paragraph{Cascaded Models.}
In Table \ref{table:multi pretrain model ablation module}, we explore how each pre-trained model contributes to the collaboration on different shots of ImageNet. Therein, ``CLIP'' denotes the zero-shot CLIP with cache model containing only CLIP's keys, and ``DINO'' denotes only the cache model with DINO's keys. As shown in the first three rows, the CLIP's language-contrastive knowledge performs stronger than DINO's vision-contrastive knowledge, which might benefit from millions of pre-training data. Their adaptive ensemble by cache model can bring larger improvement when the shot number increases. For the next two rows, DALL-E and GPT-3 can independently boost both CLIP and DINO for nearly all shots with the prompts and generated synthetic images. The last row represents our final solution, CaFo that incorporates all three pre-trained models with the best performance for all shots.

\setlength{\tabcolsep}{6pt}
    \begin{table}[t]
    \centering
    \small
    \begin{tabular}{lccccc}
    \toprule
         \multicolumn{1}{c}{\multirow{2}*{Method}}&\multicolumn{5}{c}{{Shot}}\\
         \cmidrule(lr){2-6}
         &1 &2 &4 &8 &16\\
         \cmidrule(lr){1-6}
         CLIP & 61.36 & 61.78 & 62.83 &64.04 &65.53\\
         DINO & 34.13 & 34.44 & 41.12 & 45.01 & 53.63\\
         \cmidrule(lr){1-6}
         Average &60.70 &60.72 &60.99 &61.47 &61.97\\
         Maximum &61.64 &62.45 &62.95 &63.60 &64.97\\
         \cmidrule(lr){1-6}
         $p_{\text{CLIP}}$ Base. &62.36 &63.22 &64.11 &65.50 &67.40 \\
         $p_{\text{DINO}}$ Base. &62.61 &63.39 &64.31 &65.83 &67.73\\
         $p_{\text{ZS}}$ Base. & \bf63.80 &\bf64.34 &\bf65.64 &\bf66.86 &\bf68.79\\
         \bottomrule
    \end{tabular}
    \vspace{0.05cm}
    \caption{\textbf{Ablation Study (\%) of Adaptive Inference.} We conduct different ensemble methods of cache model on ImageNet.}
    \label{tab:dynamic method ablation study}
    \end{table}
    \setlength{\tabcolsep}{1.2pt}

\paragraph{Generated Number via DALL-E.}
We utilize DALL-E to generate synthetic images as the expanded few-shot training data. In Table~\ref{tab:dalle number}, we explore the best synthetic number $K'$ for each category of different shots on ImageNet. We observe that the larger $K'$ does not lead to better few-shot performance. As we adopt pre-trained CLIP to select the top-$K'$ generated images, which are scored by the similarities between CLIP-encoded images and category texts, the larger $K'$ would contain more low-quality images and adversely affect the cache model. Furthermore, the amount of expanded data is comparable to the original $K$ shots and thus preserves the characteristic of few-shot learning. 

\paragraph{Adaptive Inference.}
In Table~\ref{tab:dynamic method ablation study}, we ablate different ensemble methods of CLIP and DINO's predictions during inference on ImageNet. The first two rows represent the cache model with one type of keys respectively for two pre-trained models without ensemble. Then, we adopt average and maximum pooling between the two predictions and ensemble the result with $p_{\text{ZS}}$. However, such naive integration without adaptive weights causes accuracy degradation. In the last three rows, we calculate the distribution similarities for adaptive ensemble and respectively select the three logits as the baseline. As shown, using $p_{\text{ZS}}$ as the distribution baseline performs the best, since $p_{\text{ZS}}$ itself shows strong transfer ability and can effectively suppress the wrong predictions of other logits.

\begin{figure}[t]
\centering
\includegraphics[width=0.93\linewidth]{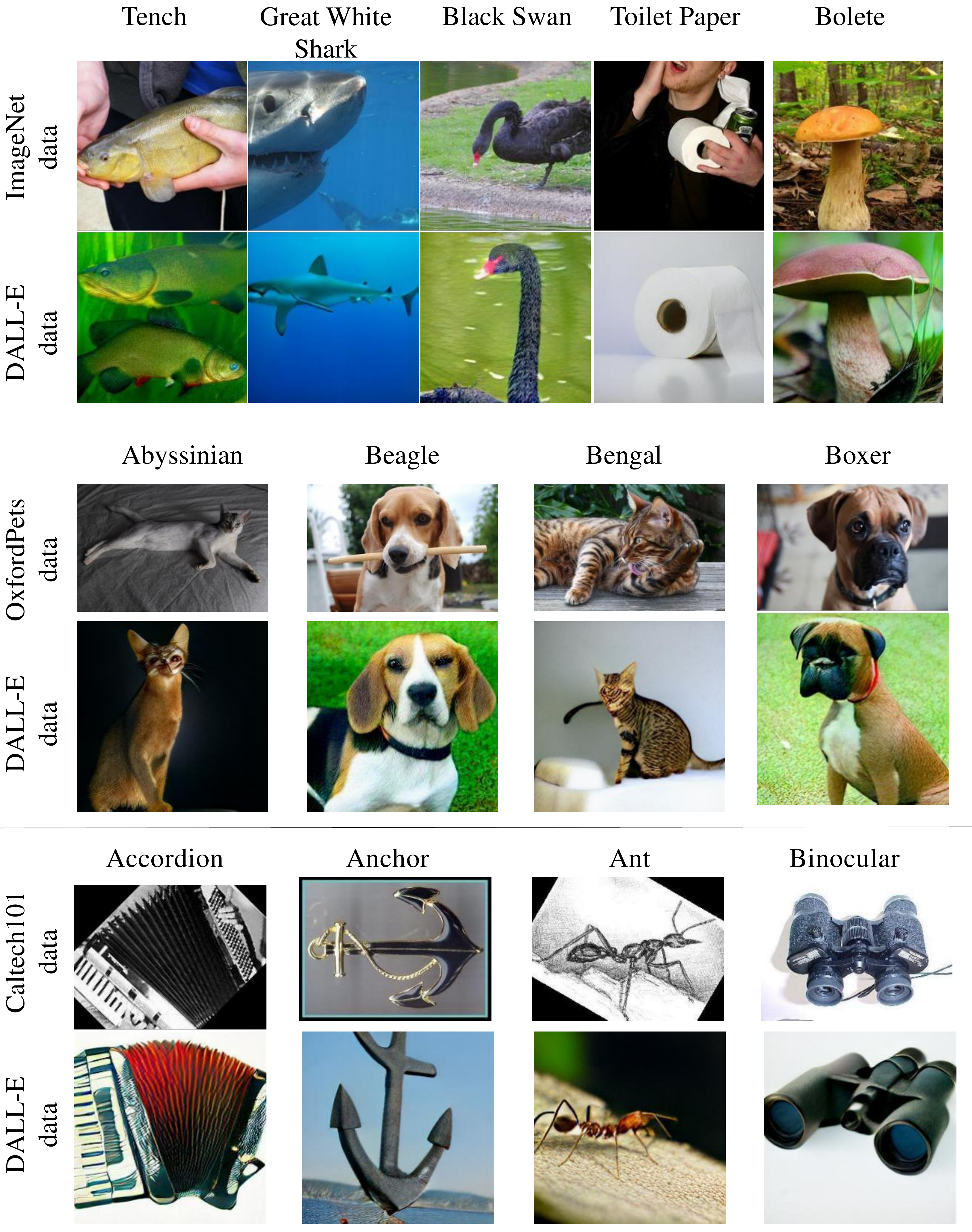}
\caption{\textbf{Visualizations of DALL-E's Generated Images.} Examples are from ImageNet, OxfordPets and Caltech101 datasets.}
\label{fig:image_comparison}
\end{figure}

\vspace{-2pt}
\setlength{\tabcolsep}{7pt}
\begin{table}[t]
\small
\begin{tabular}{lccccc}
\toprule
DALL-E & 1              & 2              & 4              & 8              & 16             \\ \midrule
1                          & 63.29          & 64.06          & 65.11          & 66.48          & 68.64          \\
2                          & 63.66          & \textbf{64.34} & 65.37          & \textbf{66.86}          & \textbf{68.79}          \\
4                          & 63.71 & 64.33          & 65.35          & 66.75          & 68.61 \\
8                          &\textbf{63.80}          & 64.26          & \textbf{65.64} & 66.68          & 68.76          \\
16                         & 63.68          & 64.16          & 65.40          & 66.57 & 68.41          \\ \bottomrule
\end{tabular}
\caption{\textbf{Ablation Study (\%) of Generated Number via DALL-E.} We compare different shot numbers on ImageNet.}
\label{tab:dalle number}
\end{table}
\vspace{0.3cm}
\setlength{\tabcolsep}{1.4pt}

\vspace{-8pt}
\paragraph{CLIP's Visual Encoders.}
We conduct CaFo with different CLIP's visual encoders for comparison with other methods. As shown in Table~\ref{tab:Different vision backbone comparison}, CaFo consistently achieves leading performance with different visual backbones, indicating our generalizability to network architectures.

\begin{figure}[t]
\centering
\includegraphics[width=0.93\linewidth]{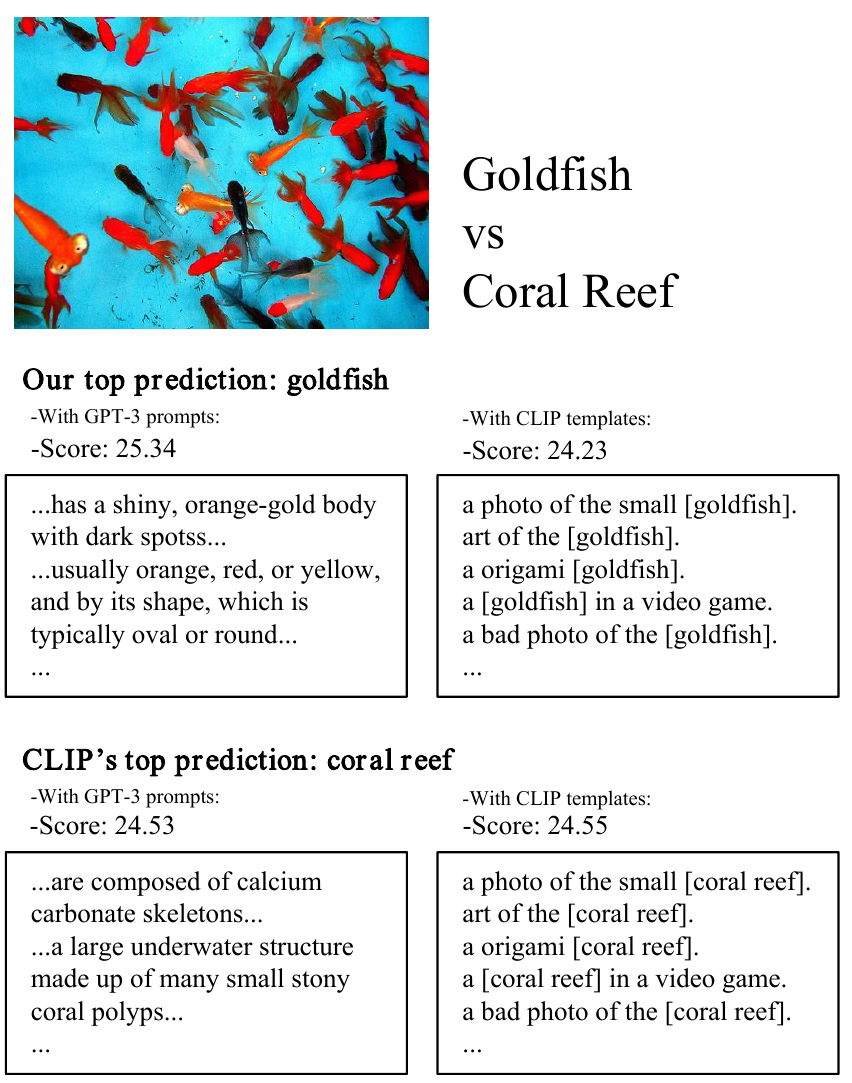}
\caption{\textbf{Visualization of GPT-3's Prompts for CLIP.} The example shown is from the ImageNet dataset.}
\label{fig:gpt_vis}
\vspace{0.25cm}
\end{figure}


\setlength{\tabcolsep}{5pt}
    \begin{table}[t!]
    \vspace{-7pt}
    \centering
    \small
    \begin{tabular}{lcccc}
    \toprule
        \multirow{1}*{Models} &RN50 &RN101 &ViT-B/32 &ViT-B/16 \\
        \cmidrule(lr){1-1}
        \cmidrule(lr){2-5}
         Zero-shot CLIP &60.33 &62.53 &63.80 &68.73\\
         CoOp &62.95 &66.60 &66.85 &71.92 \\
         CLIP-Adapter &63.59 &65.39 &66.19 &71.13 \\
         Tip-Adapter-F&65.51 &68.56 &68.65 &73.69 \\
         \textbf{CaFo} &\bf68.79 &\bf70.86 &\bf70.82 &\bf74.48 \\
         \bottomrule
    \end{tabular}
    \caption{\textbf{Ablation Study (\%) of CLIP's Visual Encoders}. We experiment different visual backbones on the 16-shot ImageNet.}
    \label{tab:Different vision backbone comparison}
    \end{table}
    \setlength{\tabcolsep}{1.4pt}

\subsection{Visualization}

\paragraph{DALL-E's Generated Images.} In Figure \ref{fig:image_comparison}, we visualize the synthetic images generated by DALL-E on ImageNet~\cite{deng2009imagenet}, OxfordPets~\cite{parkhi2012cats} and Caltech101~\cite{fei2004learning}. As shown, benefited from the vision-generative knowledge, the generated images can well highlight the downstream semantics of target category and effectively expand the few-shot training set in low-data regimes.


\paragraph{GPT-3's Prompts for CLIP.}
In Figure \ref{fig:gpt_vis}, We present a rectified example in ImageNet~\cite{deng2009imagenet} aided by GPT-3's prompts in CaFo. As shown, prompting by GPT-3 (Left) produces more semantic texts compared to CLIP's handcrafted templates(Right), and better depicts the visual appearances in the image, which predicts the correct category of goldfish.

\section{Conclusion}
We propose CaFo, a cascade of foundation models that comprehends diverse knowledge from different pre-training and follows the `Prompt, Generate, then Cache' pipeline. 
We first incorporate the generative language model, GPT-3, for prompting CLIP with more semantic texts, and adopt DALL-E to expand the few-shot training data. Then, we adaptively fuse the vision-contrastive DINO with CLIP via a unified cache model. By collaboration, CaFo achieves \textit{state-of-the-art} performance for few-shot learning on 11 datasets. Although CaFo has unified four types of pre-training, our future direction will focus on integrating more existing pre-trained knowledge, such as the masked-generative MAE~\cite{He_2022_CVPR}, the 3D-contrastive CrossPoint~\cite{afham2022crosspoint}, and 3D-generative I2P-MAE~\cite{i2p}.

{\small
\bibliographystyle{ieee_fullname}
\bibliography{aaai22}
}

\clearpage
\appendix

\section{Additional Performance Comparison}
\begin{figure*}[t!]
    \centering
    \subfloat{\includegraphics[scale=0.225]{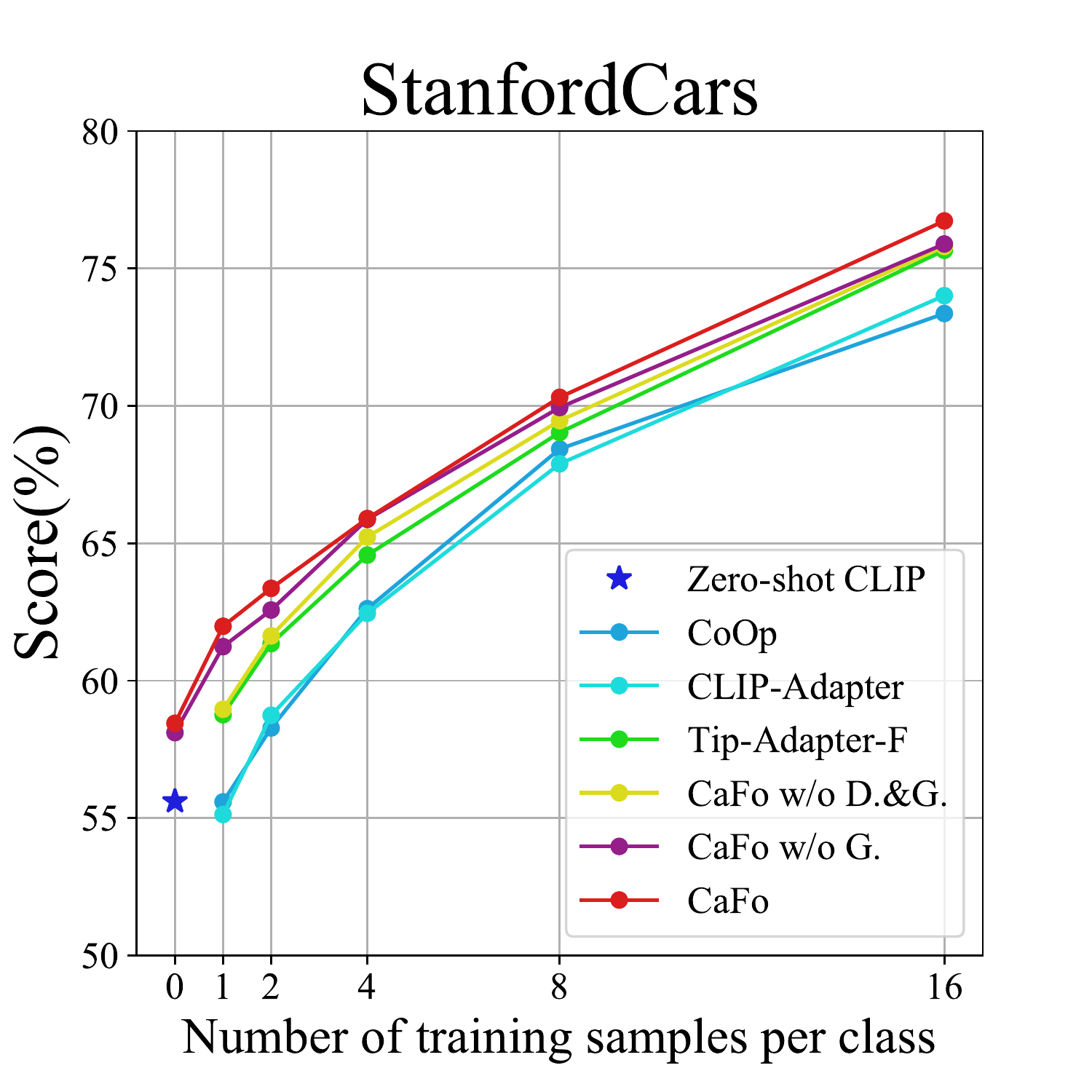}}
    \subfloat{\includegraphics[scale=0.225]{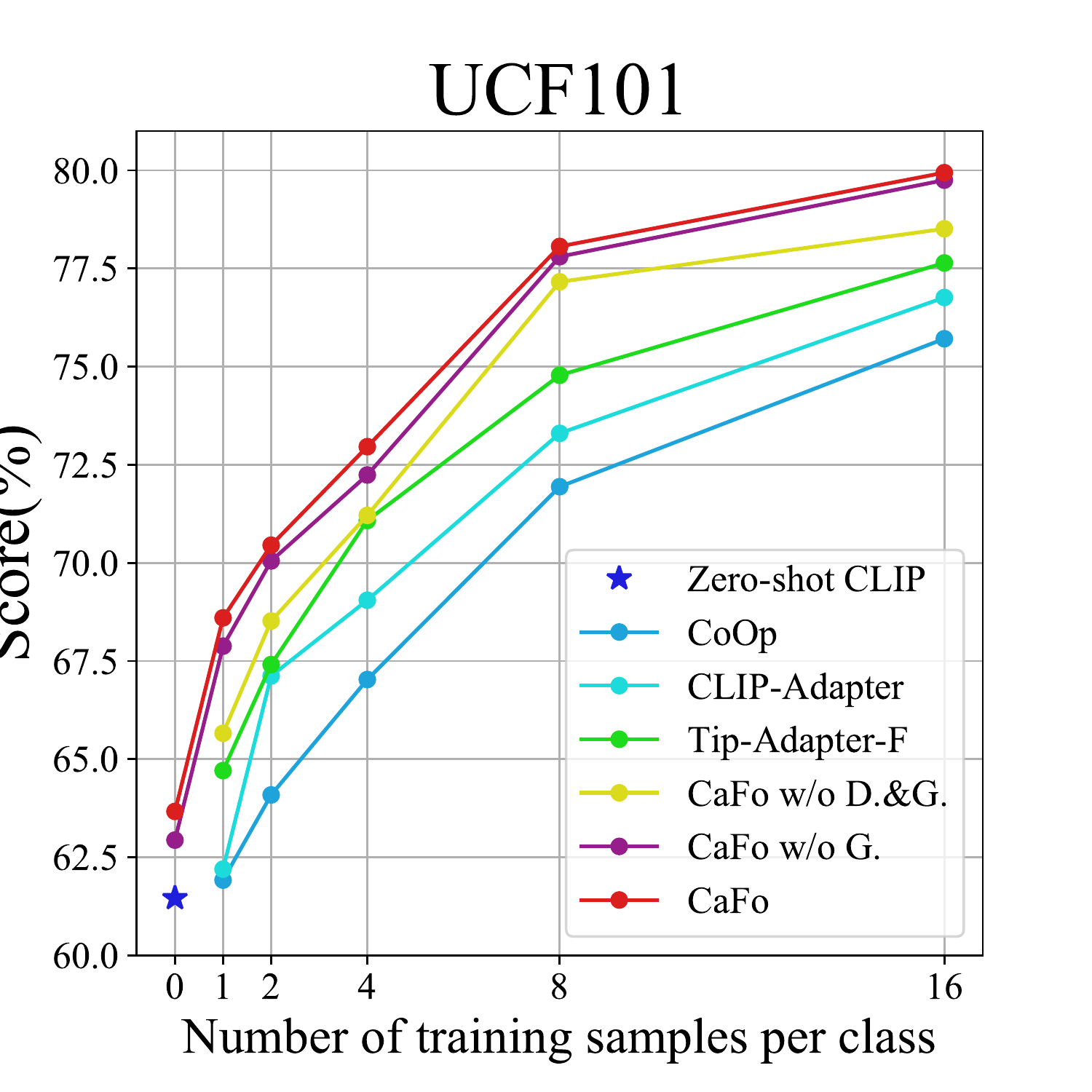}}
    \subfloat{\includegraphics[scale=0.225]{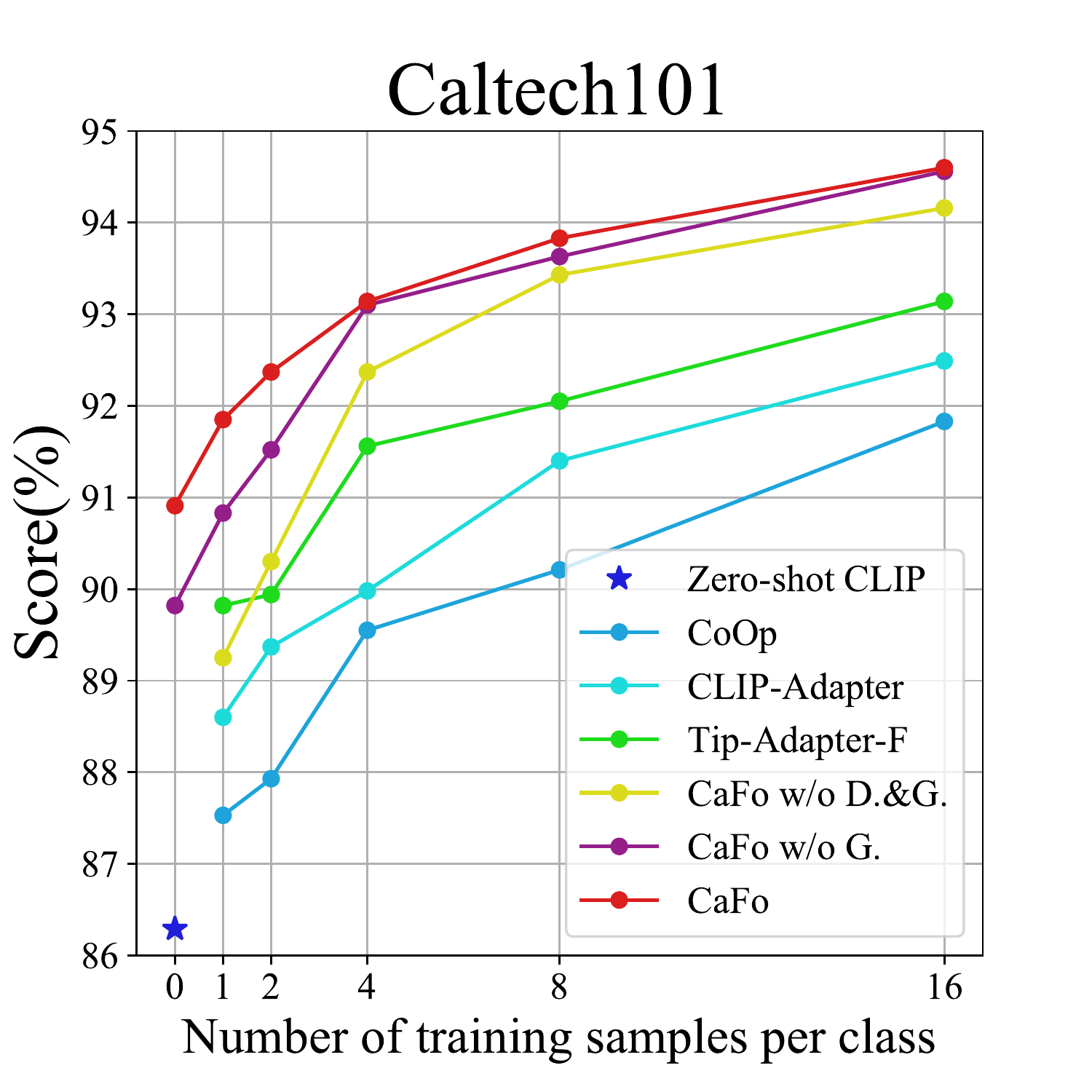}}
    \subfloat{\includegraphics[scale=0.225]{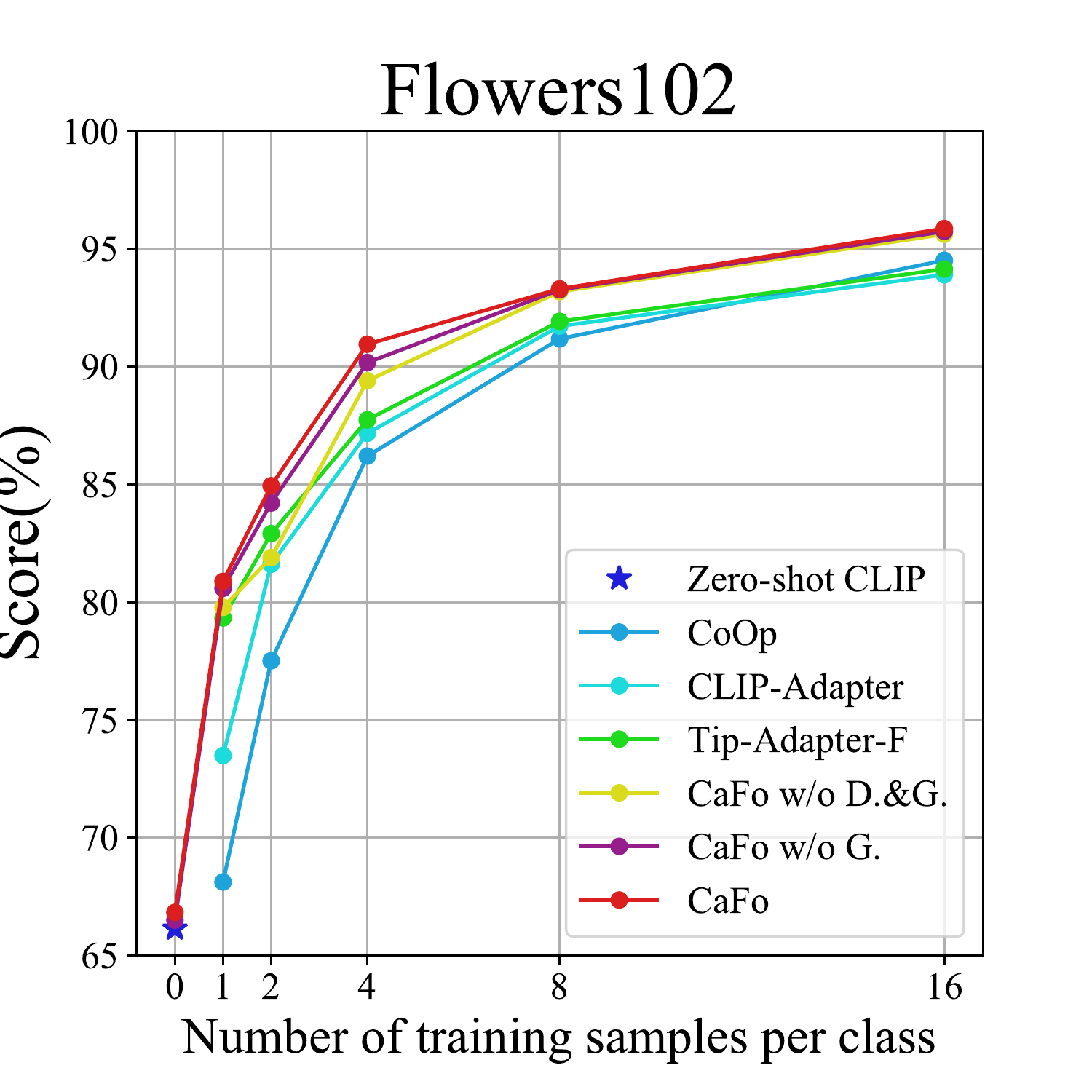}}
    \subfloat{\includegraphics[scale=0.225]{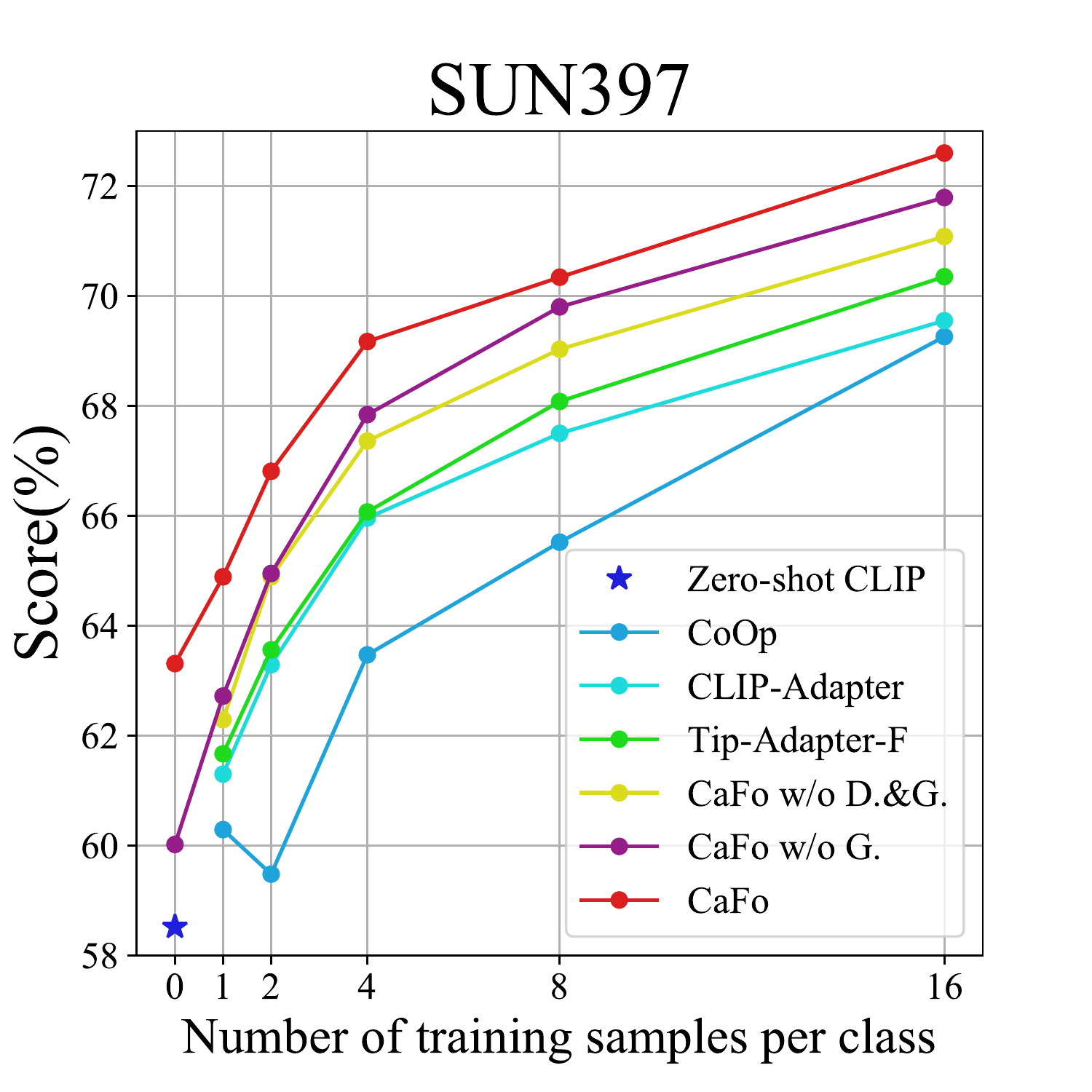}}
    \\
    \subfloat{\includegraphics[scale=0.225]{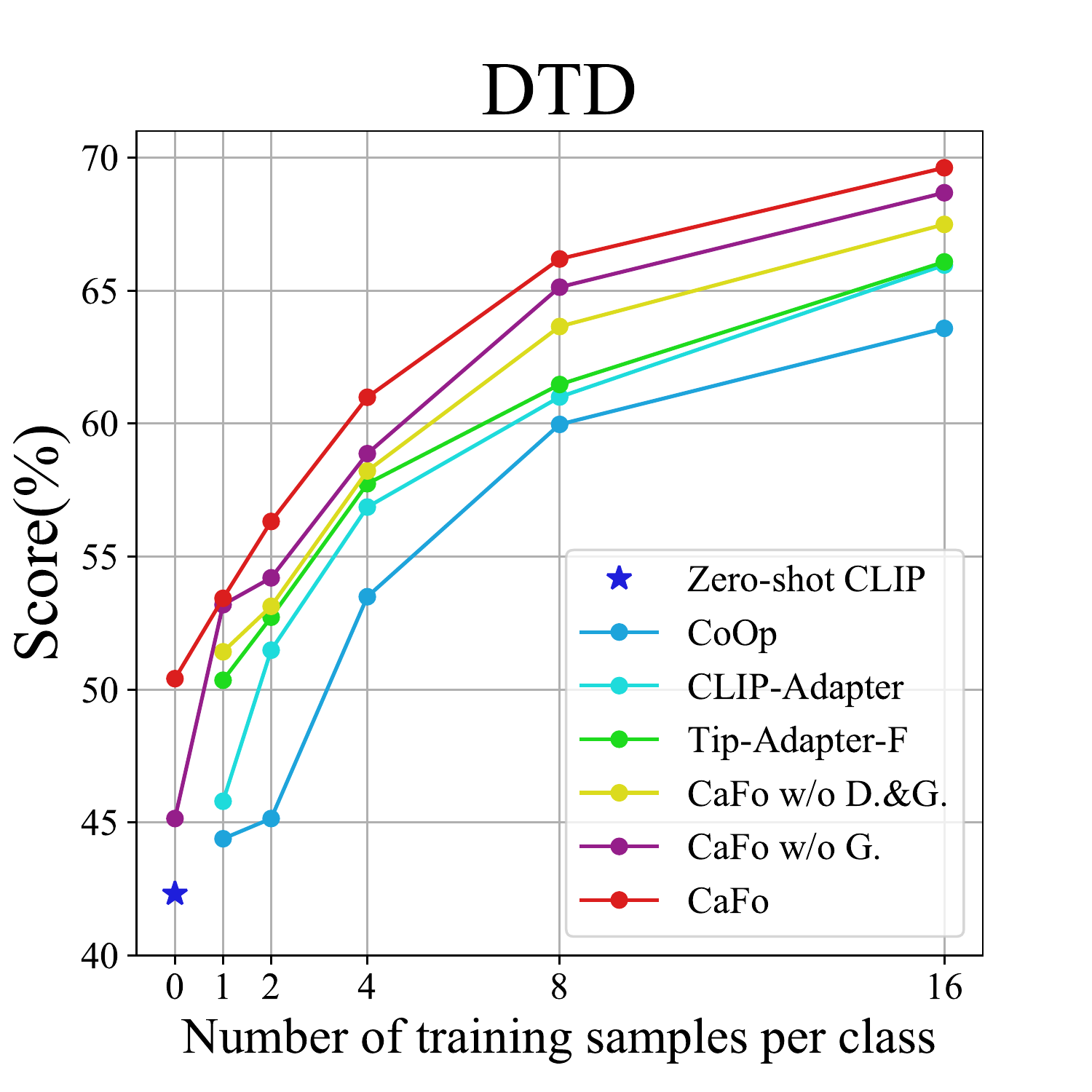}}
    \subfloat{\includegraphics[scale=0.225]{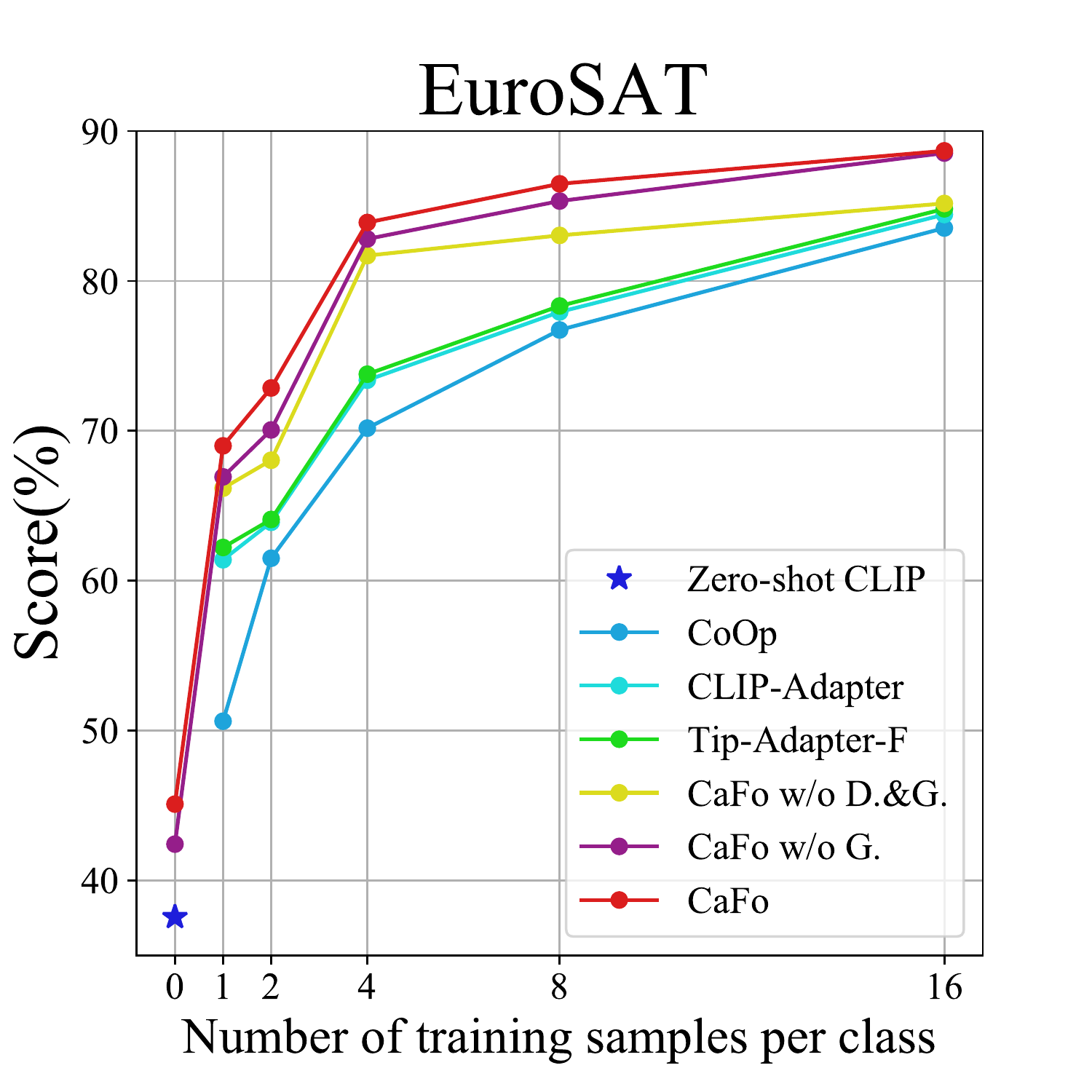}}
    \subfloat{\includegraphics[scale=0.225]{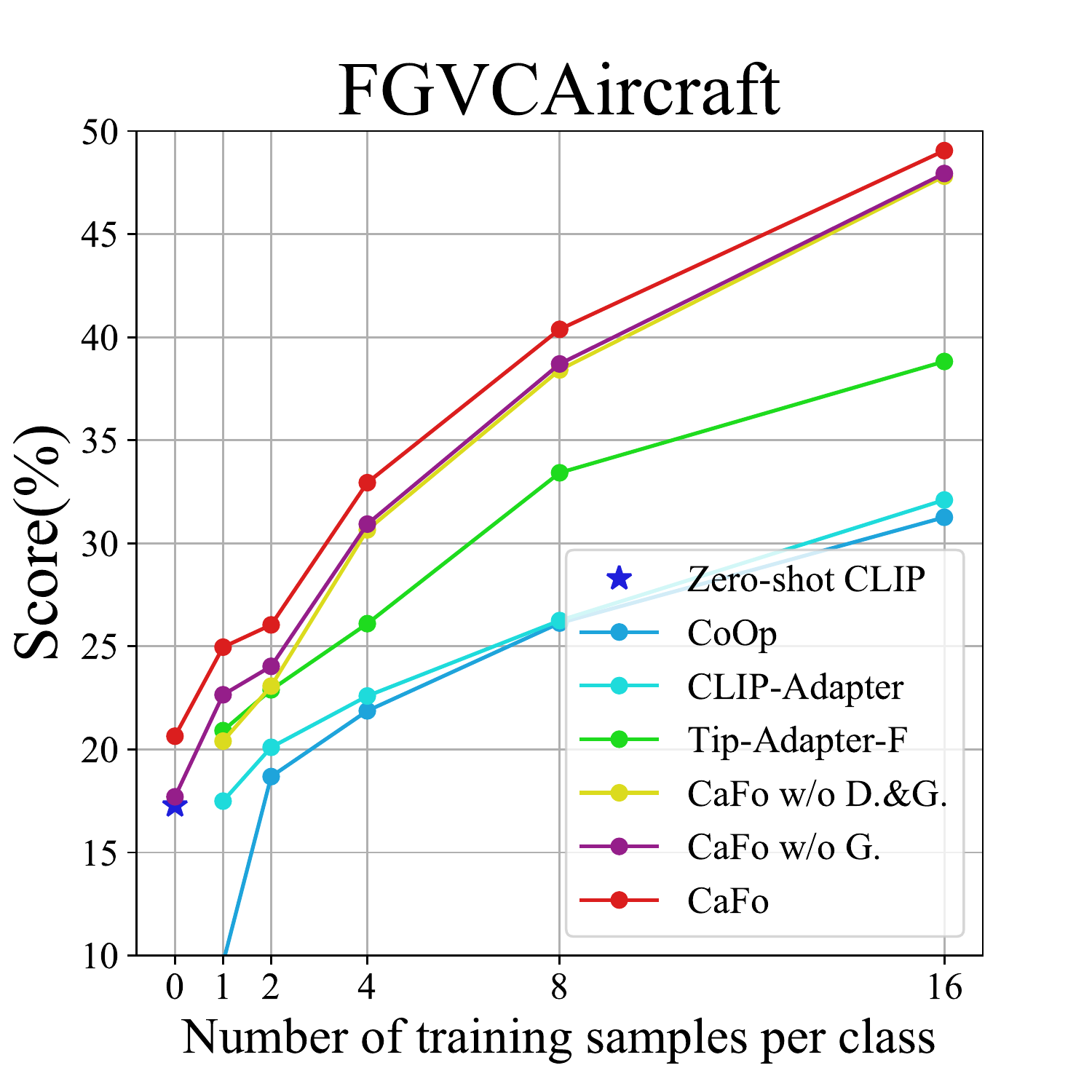}}
    \subfloat{\includegraphics[scale=0.225]{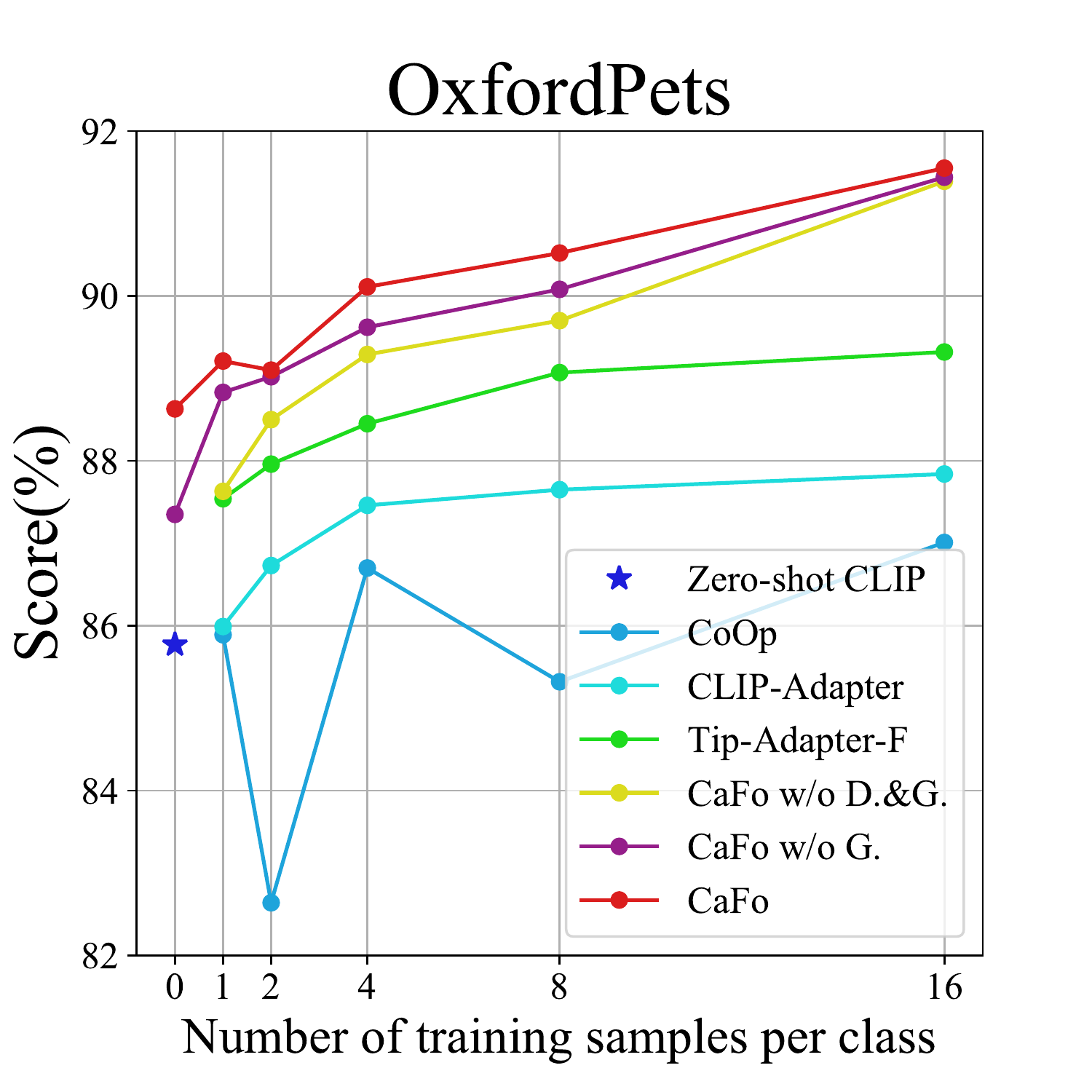}}
    \subfloat{\includegraphics[scale=0.225]{Fig/performance/Food_performance_fig.pdf}}
    \caption{\textbf{Performance (\%) Comparison on 10 Datasets.} Our method shows \textit{state-of-the-art} performance for all few-shot settings on different datasets. `CaFo w/o D.\&G.' denotes CaFo without DALL-E's generated images and GPT3's created prompts.}
    \label{fig:other_dataset_2}
    \vspace{0.3cm}
\end{figure*}

In Figure \ref{fig:other_dataset_2}, we compare the performance of CaFo without DALL-E~\cite{pmlr-v139-ramesh21a}'s generated images or GPT-3~\cite{brown2020language}'s created prompts on 10 datasets, which still consistently outperform the second-best Tip-Adapter-F.

\section{Additional Ablation Study}

\paragraph{Other Foundation Models.}
For the cache model, we investigate other pre-trained foundation models besides CLIP~\cite{radford2021learning} and DINO~\cite{Caron_2021_ICCV}, including SimCLR~\cite{chen2020improved}, MAE~\cite{he2021masked}, and SLIP~\cite{mu2022slip}. We preserve the prompting and generation by GPT-3~\cite{brown2020language} and DALL-E~\cite{pmlr-v139-ramesh21a}, along with We the $p_{\text{ZS}}$ as the ensemble baseline during adaptive inference. As shown in Table~\ref{t2222}, `CLIP+DINO', as our final solution, performs the best. Also, as an enhanced version of CLIP, SLIP can achieve higher accuracy in CaFo.

\begin{table}[h]
\centering
\small
 \begin{tabular}{ c c c c c c c}
 \toprule
 Setting &\multicolumn{2}{c}{\ ImageNet\ \ \ } &\multicolumn{2}{c}{OxfordPets} &\multicolumn{2}{c}{\ EuroSAT\ \ }\\
 \cmidrule(lr){1-1} \cmidrule(lr){2-3} \cmidrule(lr){4-5} \cmidrule(lr){6-7}
        \ \ \ \ \ CLIP+SimCLR\ \ \ \ \  &62.3 &65.7 &87.1 &89.4 &55.7 &75.9\\
        CLIP+MAE &62.2 &65.5 &87.1 &89.1 &63.7 &72.7\\
        DINO+MAE &63.0 &68.4 &88.8 &91.9 &60.0  &88.0\\
        DINO+SimCLR &63.1  &68.5& 88.8 & 91.3 &70.7  &87.7\\
 CLIP+DINO &63.8 &68.8 &89.2 &91.6 &69.0 &88.7\\
        \color{gray}{SLIP+DINO} &\color{gray}{71.0}  &\color{gray}{75.6} &\color{gray}{92.2}  &\color{gray}{94.0} &\color{gray}{71.3}  &\color{gray}{88.6}\\   
    \bottomrule
 \end{tabular}
\caption{\textbf{Ablation Study (\%) of Other Foundation Models in the Cache Model.} We report the accuracy of 1 and 16 shots on ImageNet~\cite{deng2009imagenet}, OxfordPets~\cite{parkhi2012cats}, and EuroSAT~\cite{helber2019eurosat}.}
\vspace{-0.3cm}
\label{t2222}
\end{table}

\paragraph{Zero-shot CaFo.}
As we leverage the pre-trained DALL-E to generate the supplementary few-shot training set in a zero-shot manner, our CaFo can be evaluated under zero-shot settings the same as CLIP, for which none of the human-annotated training images is given. In Table~\ref{tab:Zero shot ablation study}, we report the best generated image number $K'$ of DALL-E for zero-shot CaFo. The number ``0'' denotes Zero-shot CLIP. For different datasets, the best number varies ranging from 1$\sim$16, and the larger number normally cannot get the better result, probably due to the low-quality synthetic images. On Caltech101~\cite{fei2004learning} and EuroSAT~\cite{helber2019eurosat}, zero-shot CaFo largely surpasses CLIP by +4.62\% and +7.54\%, indicating our superiority under zero-shot settings.

\paragraph{Hyperparameter $\beta$.}
In Formula 5 and 6, we utilize a non-linear modulator $\varphi(x) = \exp(-\beta\cdot(1-x))$ for the affinity matrix of CLIP and DINO in the cache model, where $\beta$ controls the matrix sharpness. In Table~\ref{tab:hyperparameter ablation}, we experiment CaFo with different $\beta$ on 16-shot ImageNet and observe 0.6 performs the best.

\setlength{\tabcolsep}{1pt}

    \begin{table}[h]
    \begin{center}
    \small
    \begin{tabular}{c|cccccc}
    \toprule
    \text{Sharpness $\beta$\ } & 0.4 & 0.5 & 0.6 & 0.7 & 0.8 & 1.0 \\
    \cmidrule(lr){1-1} \cmidrule(lr){2-7}
     CaFo &\ \ 68.66\ \ &\ \ 68.75\ \ &\ \ \bf68.79\ \ &\ \ 68.73\ \ &\ \ 68.69\ \ &\ \ 68.66\ \ \\
    \bottomrule
    \end{tabular}
    \end{center}
    \vspace{-8pt}
    \caption{\textbf{Ablation Study (\%) of Hyperparameter $\beta$.} We report the 16-shot accuracy on ImageNet~\cite{deng2009imagenet}.}
    \label{tab:hyperparameter ablation}
    \end{table}
    \vspace{-8pt}
    \setlength{\tabcolsep}{1pt}

\setlength{\tabcolsep}{9pt}
    \begin{table*}[t]
    \centering
    \begin{tabular}{cccccccccccc}
    \toprule
        {DALL-E} & \rotatebox{90}{ImageNet} & \rotatebox{90}{Caltech101} & \rotatebox{90}{Flower102} & \rotatebox{90}{Food101} & \rotatebox{90}{DTD} & \rotatebox{90}{EuroSAT} & \rotatebox{90}{OxfordPets} & \rotatebox{90}{SUN397} & \rotatebox{90}{StanfordCars} & \rotatebox{90}{UCF101} & \rotatebox{90}{FGVCAircraft}\\
        \cmidrule(lr){1-1}
        \cmidrule(lr){2-12}
        0 & 60.33 & 86.29 & 66.14 & 77.20 & 50.30 & 37.56 & 85.77 & 58.52 & 55.61 & 61.46 & 17.28\\
        1 & 62.5 & 89.78 & 65.65 & 77.52 & 50.12 & 37.46 & 87.33 & 63.08 & 57.33 & 63.05 & 20.46\\
        2 & 62.69 & 90.26 & \bf66.83 & 77.50 & 50.00 & 41.73 & 87.49 & 63.02 & 57.63 & 62.44 & 20.31\\
        4 & 62.81 & 89.98 & 66.50 & \bf77.58 & \bf50.41 & 43.2 & 87.71 & \bf63.31 & 57.46 & 63.12 & 20.64\\
        8 & \bf62.99 & 90.67 & 66.83 & 77.56 & 50.12 & \bf45.10 & \bf88.63 & 63.26 & 58.03 & 62.83 & 20.49\\
        16 & 62.74 & \bf90.91 & 66.54 & 77.53 & 50.24 & 42.73 & 87.49 & 63.16 & \bf58.45 & \bf63.67 & \bf21.06\\
        \bottomrule
    \end{tabular}
    \caption{\textbf{Ablation Study (\%) of Zero-shot CaFo via DALL-E on Different Datasets.} We leverage DALL-E to generate different numbers of synthetic images for zero-shot recognition.}
    \label{tab:Zero shot ablation study}
    \end{table*}
    \setlength{\tabcolsep}{1.4pt}

\section{Additional Visualization}
\paragraph{GPT-3's Prompts for CLIP.} In Figure \ref{fig:gpt3_1} and \ref{fig:gpt3_2}, we show more visualization of the prompts produced by GPT-3 and how they assist our CaFo to rectify false predictions of the original CLIP's templates.
\paragraph{DALL-E's Generated Images.} In Figure \ref{fig:dalle}, we visualize more synthetic images generated
by DALL-E on different datasets.  Benefited from the
pre-trained DALL-E, the generated images can well highlight the semantics of target category and effectively expand
the few-shot training set in low-data regimes.

\begin{figure}[h]
    \centering
    \includegraphics[width=0.7\linewidth]{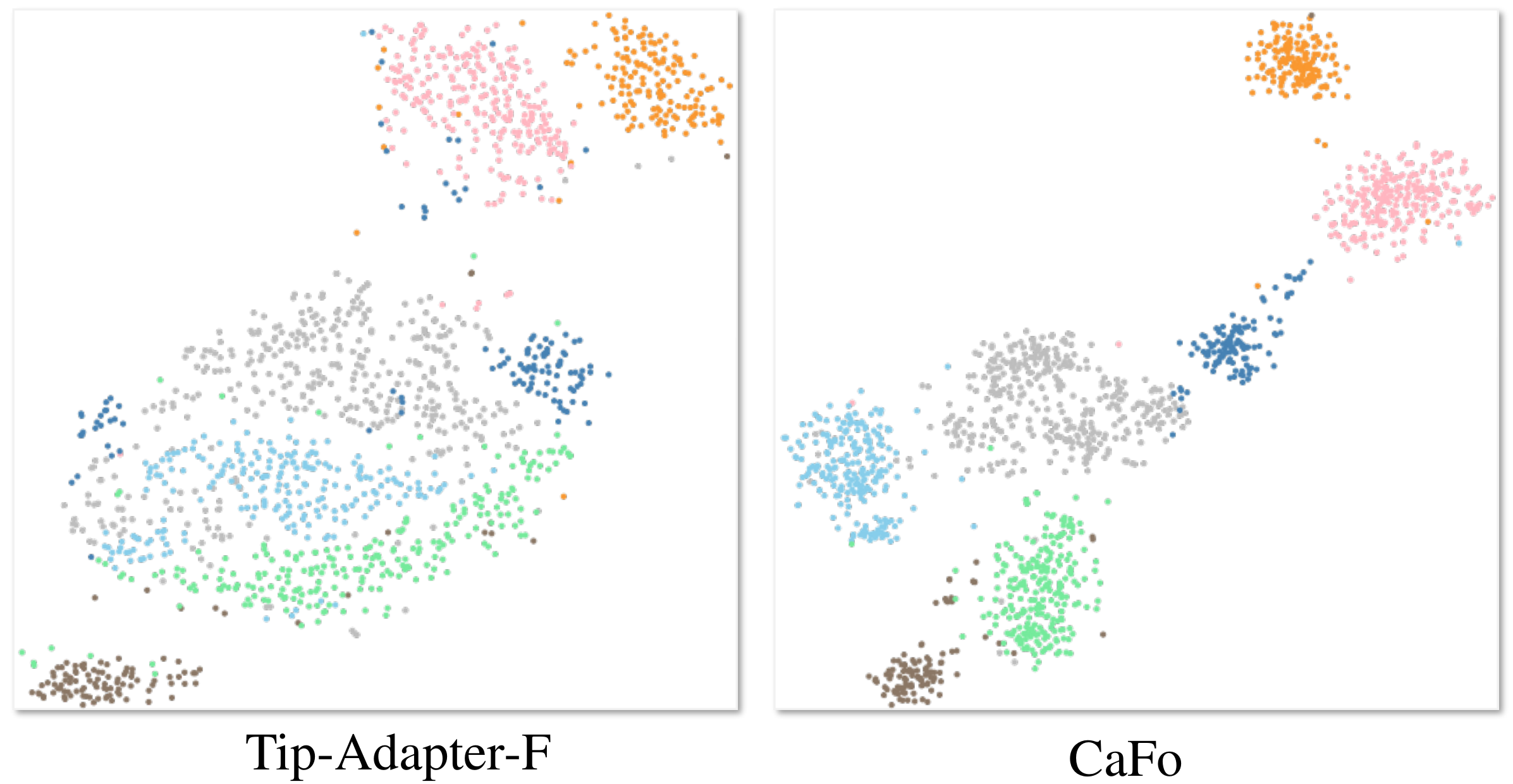}
    \caption{\textbf{t-SNE Visualization.} Different colors represent different categories on 16-shot ImageNet.}
    \label{fig:tsne}
\end{figure}

\paragraph{t-SNE.} We present the t-SNE visualization of our CaFo
and the second-best Tip-Adapter-F in Figure \ref{fig:tsne}. CaFo shows more contrastive distribution of category clusters and well mitigates some aliasing between similar classes.

\begin{figure}[h]
    \centering
    \includegraphics[width=0.77\linewidth]{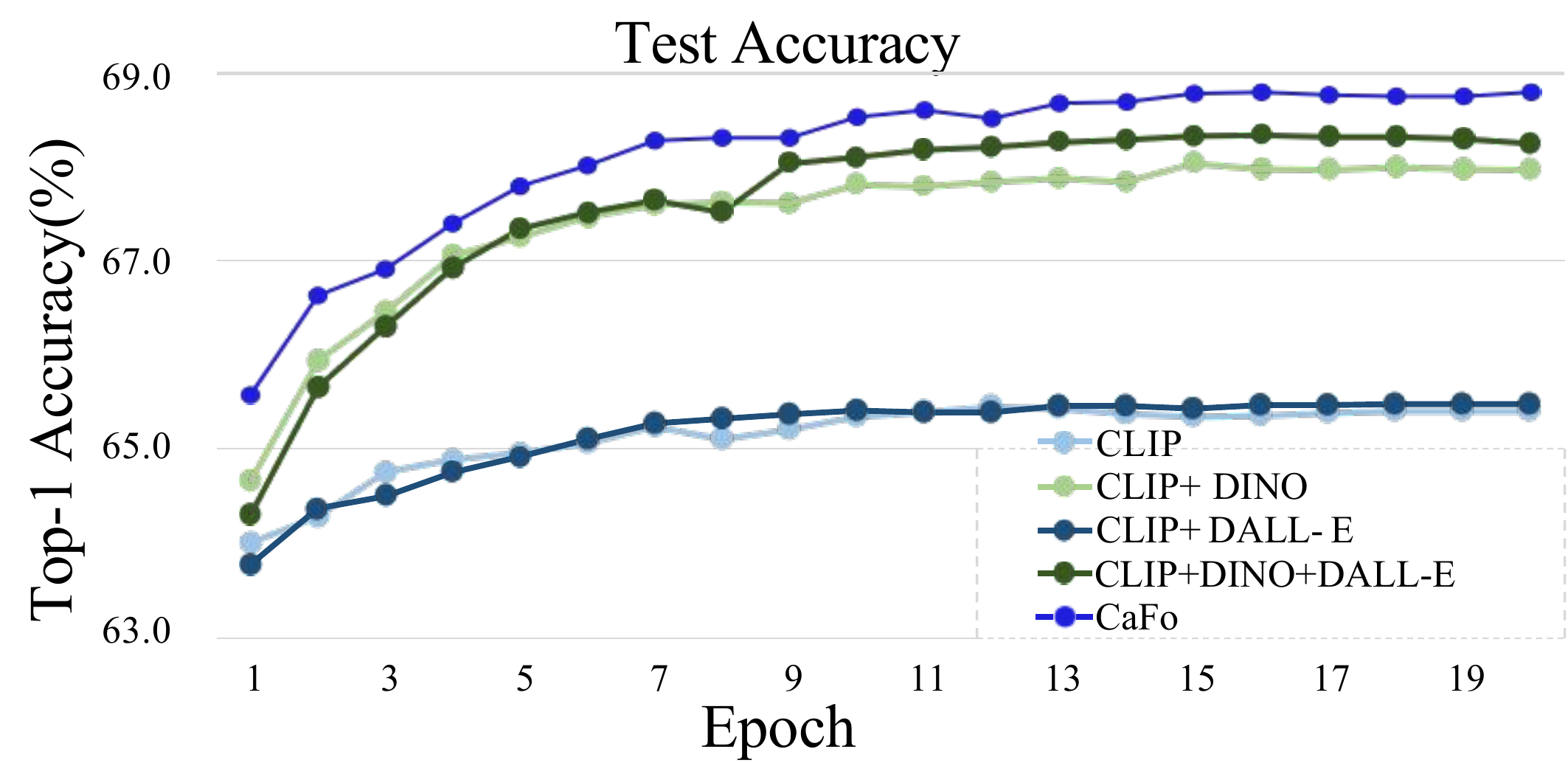}
    \caption{\textbf{Learning Curves of Test Accuracy (\%)} for different combinations of pre-trained models
 on 16-shot ImageNet.}
    \label{fig:curve}
\end{figure}

\begin{figure*}[t!]
    \centering
    \subfloat{\includegraphics[width=1\linewidth]{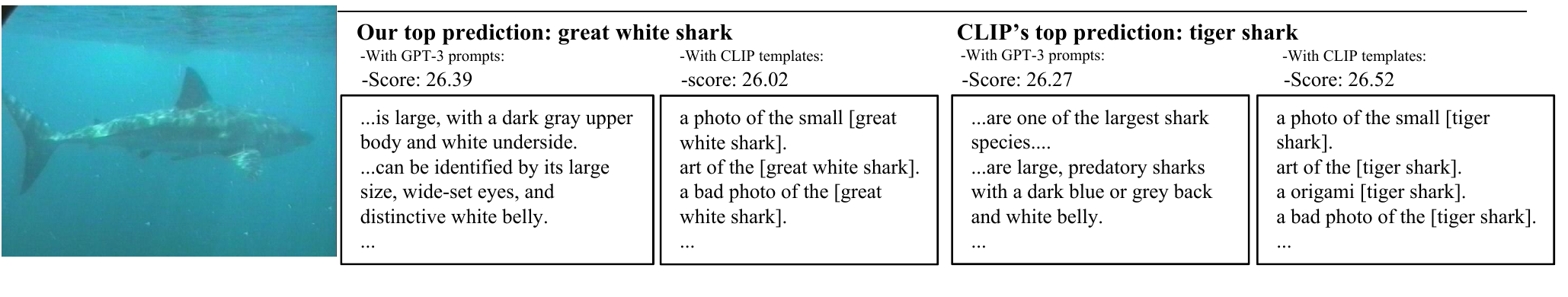}}
    \\
    \subfloat{\includegraphics[width=1\linewidth]{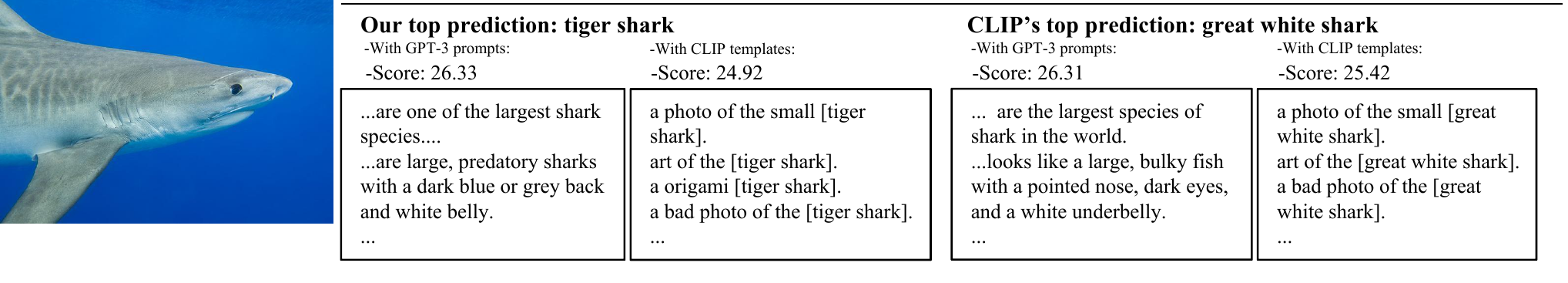}}
    \\
    \subfloat{\includegraphics[width=1\linewidth]{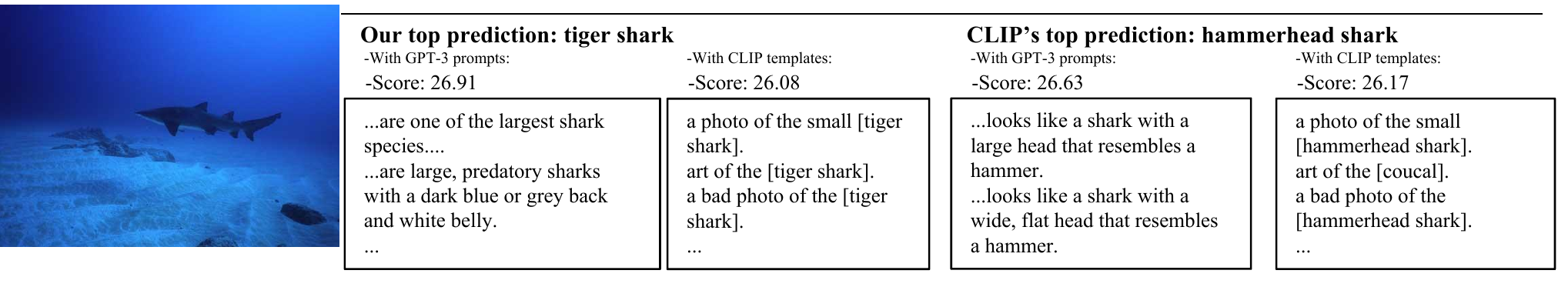}}
    \\
    \subfloat{\includegraphics[width=1\linewidth]{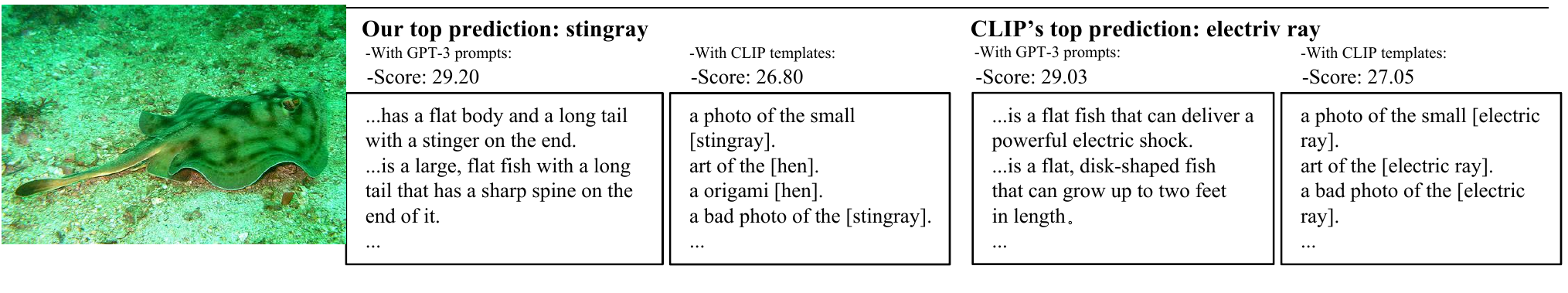}}
    
    \caption{\textbf{Additional Visualization of GPT-3's Prompts for CLIP.} Above examples are from the ImageNet dataset.}
    \label{fig:gpt3_1}
\end{figure*}

\begin{figure*}[t!]
    \centering
    \subfloat{\includegraphics[width=1\linewidth]{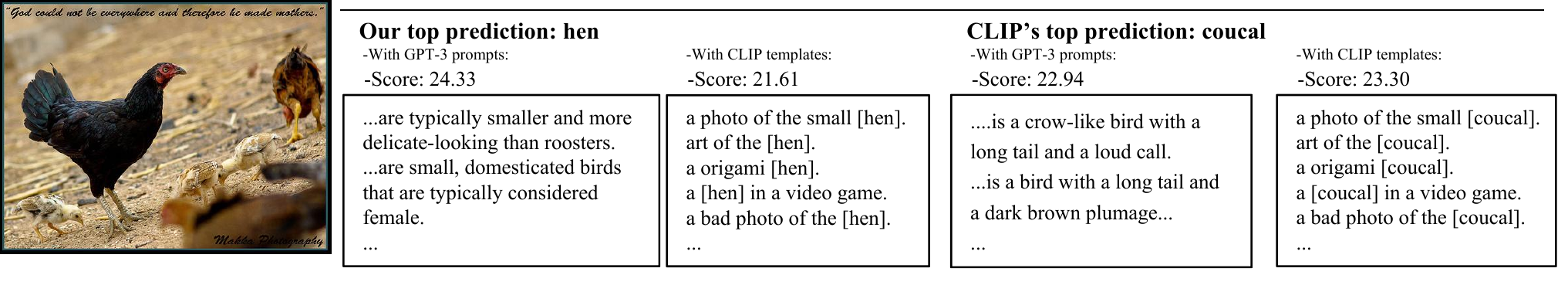}}
    \\
    \subfloat{\includegraphics[width=1\linewidth]{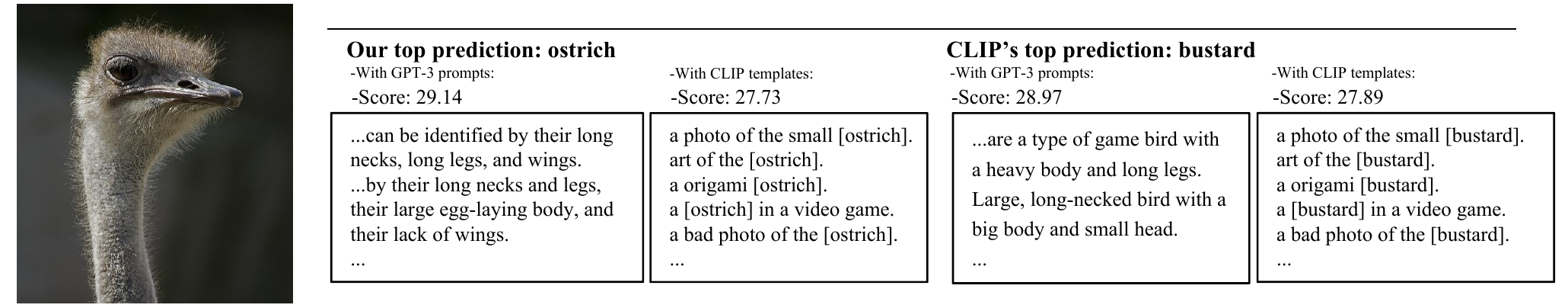}}
    \\
    \subfloat{\includegraphics[width=1\linewidth]{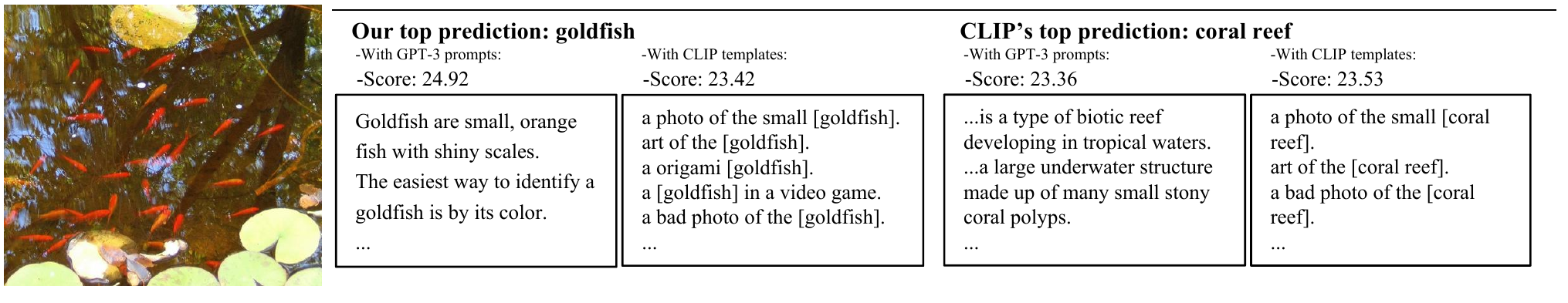}}
    \\
    \subfloat{\includegraphics[width=1\linewidth]{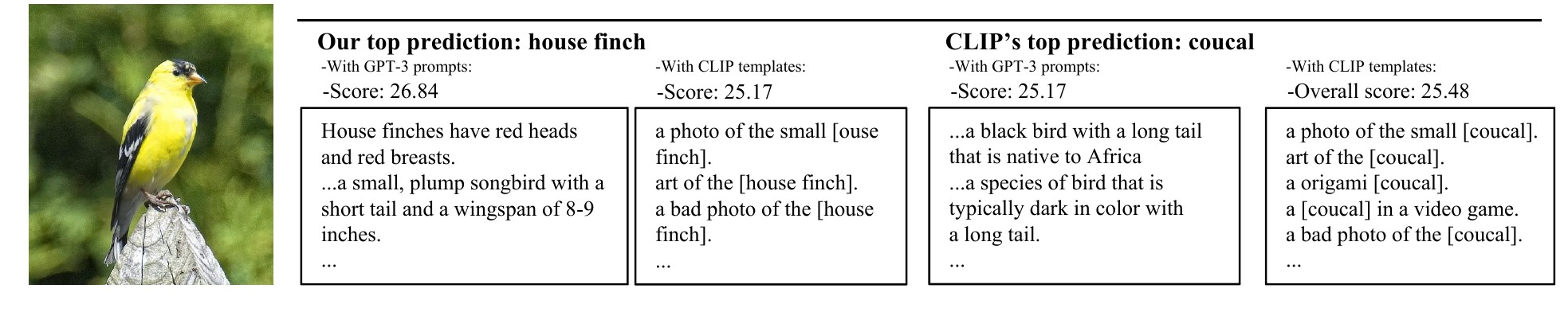}}
    \caption{\textbf{Additional Visualization of GPT-3's Prompts for CLIP.} Above examples are from the ImageNet dataset.}
    \label{fig:gpt3_2}
\end{figure*}

\begin{figure*}[t]
    \centering
    \includegraphics[width=1\linewidth]{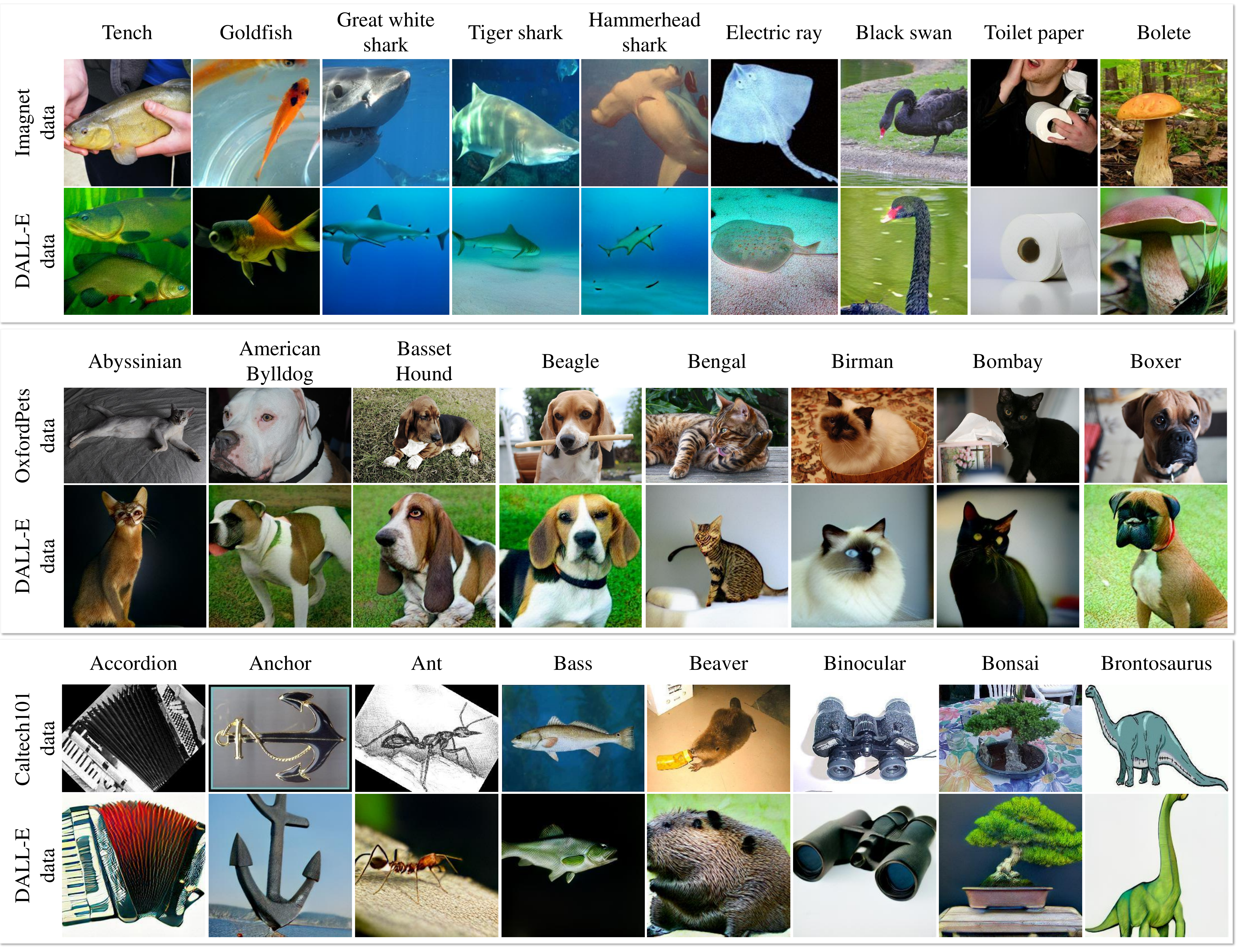}
    \caption{\textbf{Additional Visualization of DALL-E's Generated Images.} Examples are from ImageNet, OxfordPets and Caltech101 datasets.}
    \vspace{-12pt}
    \label{fig:dalle}
\end{figure*}

\paragraph{Learning Curves.}In Figure \ref{fig:curve}, we visualize the 20-epoch
learning curves of test accuracy on 16-shot ImageNet. Compared to the single CLIP, collaborating with
DALL-E, DINO and GPT-3 significantly improves the convergence speed and
classification accuracy on test set.


\end{document}